\documentclass[10pt,twocolumn,letterpaper]{article}

\usepackage{iccv}
\usepackage{times}
\usepackage{epsfig}
\usepackage{graphicx}
\usepackage{amsmath}
\usepackage{amssymb}
\usepackage{bm}       % Bold math.
\usepackage{caption}  % For first page figure hack.
\usepackage{stmaryrd} % For \varodot (component-wise product).
\usepackage{multirow} % Multicells in tables.
\usepackage{diagbox}  % Box with diagonal line and two fields in tables.
\usepackage{tikz}     % Figures.
\usetikzlibrary{calc} % Offset calculations.
\usetikzlibrary{3d}   % Cubes.
\usepackage{currfile} % Relative inputs.
\usepackage[caption=false]{subfig}  % Subfigures.
\usepackage[T1]{fontenc}  % Underscores in filenames.
\usepackage[absolute]{textpos}  % Institutions.

\usepackage[pagebackref=true,colorlinks,bookmarks=false]{hyperref}

\newcommand{\textperc}[0]{\%}
\newcommand{\chic}[0]{Chictopia10K}

\def\cn/{ClothNet}
\def\cnplain/{\cn/-full}
\def\cncond/{\cn/-body}
\def\portray/{portray}
\iccvfinalcopy % *** Uncomment this line for the final submission

\def\iccvPaperID{1783} % *** Enter the ICCV Paper ID here

\begin{document}

%%%%%%%%% TITLE
\title{A Generative Model of People in Clothing}
\author{Christoph Lassner\textsuperscript{1, 2}\\
{\tt\small classner@tue.mpg.de}
\and
Gerard Pons-Moll\textsuperscript{2}\\
{\tt\small gpons@tue.mpg.de}
\and
Peter V. Gehler\textsuperscript{3,*}\\
{\tt\small peter.gehler@uni-wuerzburg.de}
}
{\ificcvfinal}
	{\TPshowboxesfalse}
	\begin{textblock*}{\textwidth} (2cm, 6.1cm) {
      \noindent
      \begin{center}
        \textsuperscript{1}BCCN, T\"ubingen\quad
        \textsuperscript{2}MPI for Intelligent Systems, T\"ubingen\quad
        \textsuperscript{3}University of W\"urzburg
      \end{center}
    }
  \end{textblock*}
\fi

\twocolumn[{%
\renewcommand\twocolumn[1][]{#1}%
\maketitle
{\ificcvfinal}
\fi
\vspace*{2.3cm}
\hspace*{-0.5\textwidth}
\hspace*{0.07cm}
\begin{tikzpicture}[minimum height=6.000000, minimum width=\textwidth]
\node [matrix] at (0.000000, -0.000000) {\includegraphics[height=1.800000cm]{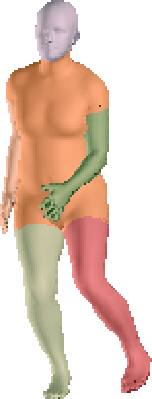}\\};
\node [matrix] at (1.400000, -0.000000) {\includegraphics[height=1.800000cm]{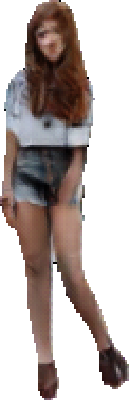}\\};
\node [matrix] at (2.400000, -0.000000) {\includegraphics[height=1.800000cm]{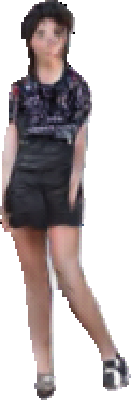}\\};
\node [matrix] at (3.400000, -0.000000) {\includegraphics[height=1.800000cm]{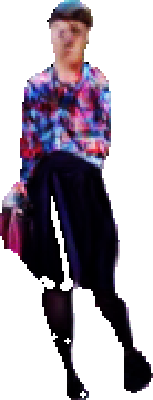}\\};
\node [matrix] at (4.400000, -0.000000) {\includegraphics[height=1.800000cm]{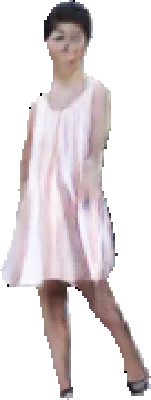}\\};
\node [matrix] at (5.400000, -0.000000) {\includegraphics[height=1.800000cm]{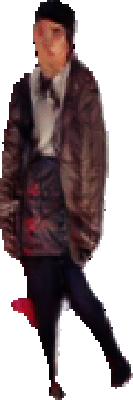}\\};
\node [matrix] at (6.400000, -0.000000) {\includegraphics[height=1.800000cm]{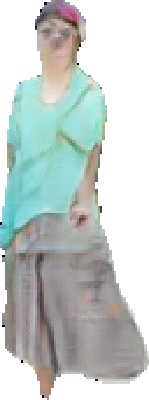}\\};
\node [matrix] at (7.400000, -0.000000) {\includegraphics[height=1.800000cm]{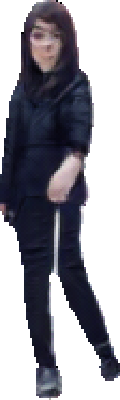}\\};
\node [matrix] at (8.400000, -0.000000) {\includegraphics[height=1.800000cm]{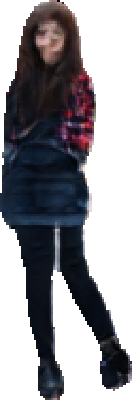}\\};
\node [matrix] at (9.400000, -0.000000) {\includegraphics[height=1.800000cm]{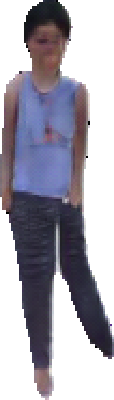}\\};
\node [matrix] at (10.400000, -0.000000) {\includegraphics[height=1.800000cm]{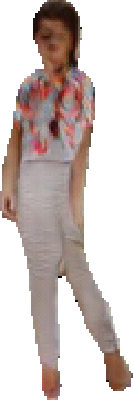}\\};
\node [matrix] at (11.400000, -0.000000) {\includegraphics[height=1.800000cm]{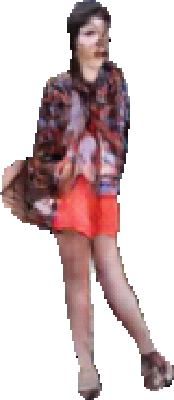}\\};
\node [matrix] at (12.400000, -0.000000) {\includegraphics[height=1.800000cm]{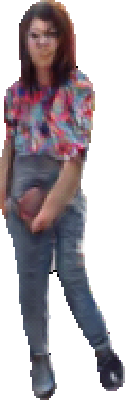}\\};
\node [matrix] at (13.400000, -0.000000) {\includegraphics[height=1.800000cm]{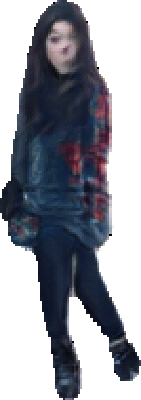}\\};
\node [matrix] at (14.400000, -0.000000) {\includegraphics[height=1.800000cm]{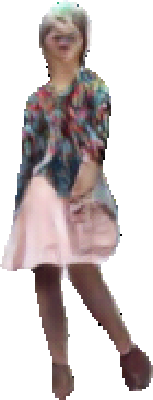}\\};
\node [matrix] at (15.400000, -0.000000) {\includegraphics[height=1.800000cm]{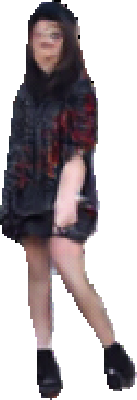}\\};
\node [matrix] at (16.400000, -0.000000) {\includegraphics[height=1.800000cm]{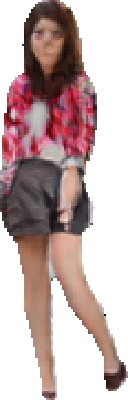}\\};
\node [matrix] at (0.500000, -1.600000) {\includegraphics[height=1.800000cm]{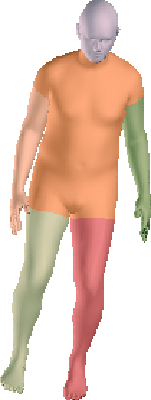}\\};
\node [matrix] at (1.900000, -1.600000) {\includegraphics[height=1.800000cm]{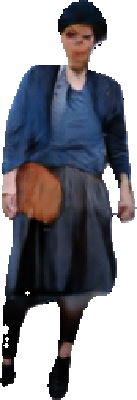}\\};
\node [matrix] at (2.900000, -1.600000) {\includegraphics[height=1.800000cm]{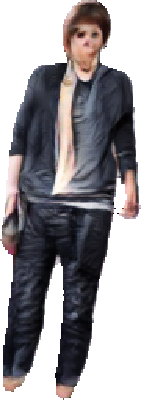}\\};
\node [matrix] at (3.900000, -1.600000) {\includegraphics[height=1.800000cm]{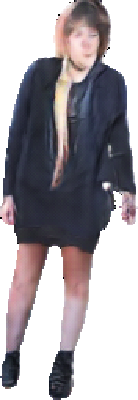}\\};
\node [matrix] at (4.900000, -1.600000) {\includegraphics[height=1.800000cm]{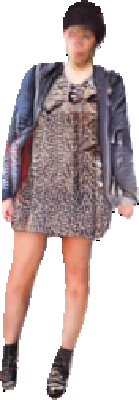}\\};
\node [matrix] at (5.900000, -1.600000) {\includegraphics[height=1.800000cm]{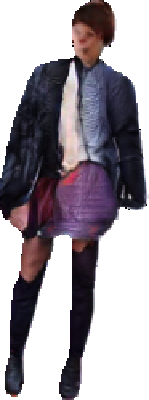}\\};
\node [matrix] at (6.900000, -1.600000) {\includegraphics[height=1.800000cm]{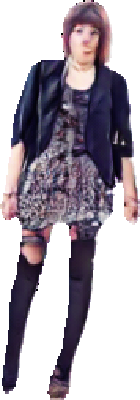}\\};
\node [matrix] at (7.900000, -1.600000) {\includegraphics[height=1.800000cm]{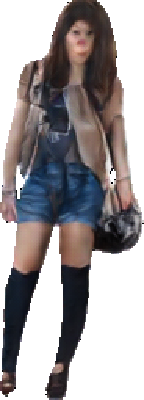}\\};
\node [matrix] at (8.900000, -1.600000) {\includegraphics[height=1.800000cm]{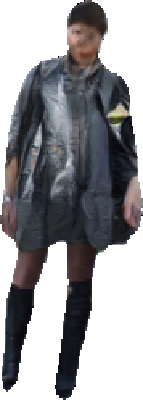}\\};
\node [matrix] at (9.900000, -1.600000) {\includegraphics[height=1.800000cm]{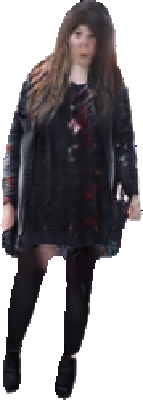}\\};
\node [matrix] at (10.900000, -1.600000) {\includegraphics[height=1.800000cm]{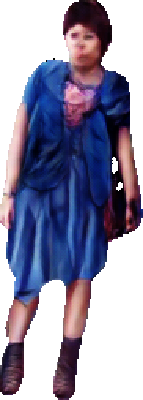}\\};
\node [matrix] at (11.900000, -1.600000) {\includegraphics[height=1.800000cm]{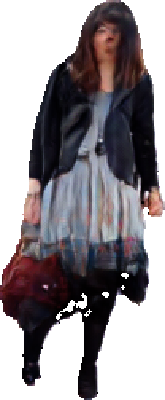}\\};
\node [matrix] at (12.900000, -1.600000) {\includegraphics[height=1.800000cm]{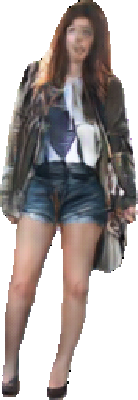}\\};
\node [matrix] at (13.900000, -1.600000) {\includegraphics[height=1.800000cm]{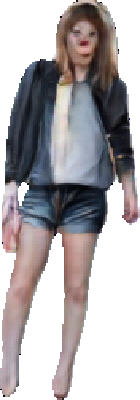}\\};
\node [matrix] at (14.900000, -1.600000) {\includegraphics[height=1.800000cm]{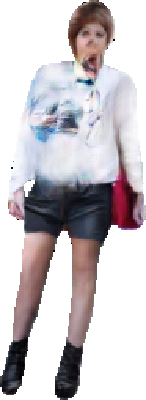}\\};
\node [matrix] at (15.900000, -1.600000) {\includegraphics[height=1.800000cm]{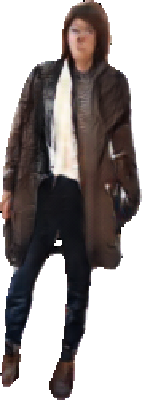}\\};
\node [matrix] at (0.000000, -3.200000) {\includegraphics[height=1.800000cm]{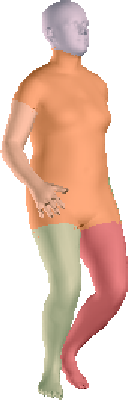}\\};
\node [matrix] at (1.400000, -3.200000) {\includegraphics[height=1.800000cm]{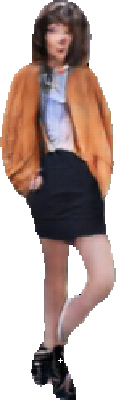}\\};
\node [matrix] at (2.400000, -3.200000) {\includegraphics[height=1.800000cm]{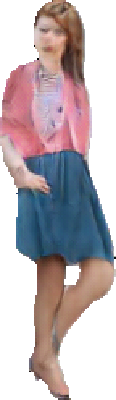}\\};
\node [matrix] at (3.400000, -3.200000) {\includegraphics[height=1.800000cm]{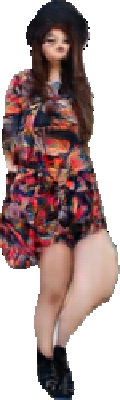}\\};
\node [matrix] at (4.400000, -3.200000) {\includegraphics[height=1.800000cm]{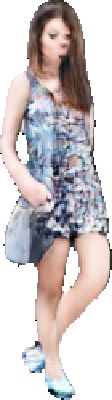}\\};
\node [matrix] at (5.400000, -3.200000) {\includegraphics[height=1.800000cm]{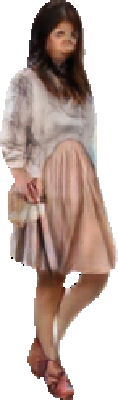}\\};
\node [matrix] at (6.400000, -3.200000) {\includegraphics[height=1.800000cm]{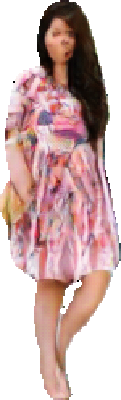}\\};
\node [matrix] at (7.400000, -3.200000) {\includegraphics[height=1.800000cm]{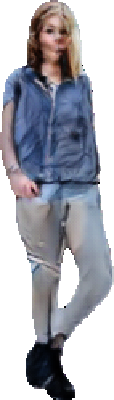}\\};
\node [matrix] at (8.400000, -3.200000) {\includegraphics[height=1.800000cm]{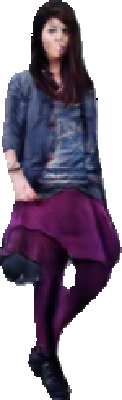}\\};
\node [matrix] at (9.400000, -3.200000) {\includegraphics[height=1.800000cm]{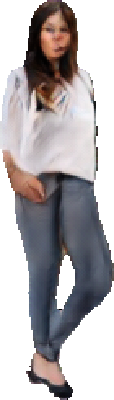}\\};
\node [matrix] at (10.400000, -3.200000) {\includegraphics[height=1.800000cm]{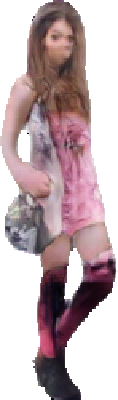}\\};
\node [matrix] at (11.400000, -3.200000) {\includegraphics[height=1.800000cm]{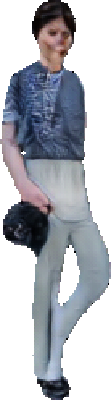}\\};
\node [matrix] at (12.400000, -3.200000) {\includegraphics[height=1.800000cm]{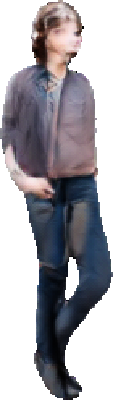}\\};
\node [matrix] at (13.400000, -3.200000) {\includegraphics[height=1.800000cm]{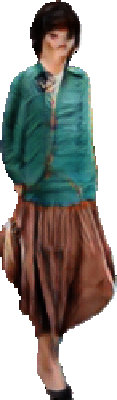}\\};
\node [matrix] at (14.400000, -3.200000) {\includegraphics[height=1.800000cm]{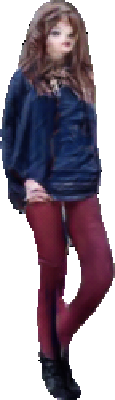}\\};
\node [matrix] at (15.400000, -3.200000) {\includegraphics[height=1.800000cm]{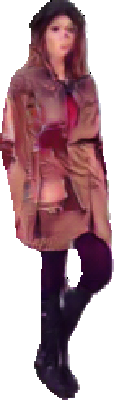}\\};
\node [matrix] at (16.400000, -3.200000) {\includegraphics[height=1.800000cm]{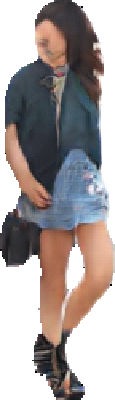}\\};
\end{tikzpicture}

\vspace*{-0.9cm}
\begin{center}
  \captionof{figure}{
    Random examples of people generated with our model. For each row, sampling is conditioned on the silhouette displayed on the left. Our proposed framework also supports unconditioned sampling as well as conditioning on local appearance cues, such as color.}
  \label{fig:teaser}
\end{center}
}]

%%%%%%%%% ABSTRACT
\begin{abstract}
  We present the first image-based generative model of people in clothing for the full body. We sidestep the commonly used complex graphics rendering pipeline and the need for high-quality 3D scans of dressed people. Instead, we learn generative models from a large image database. The main challenge is to cope with the high variance in human pose, shape and appearance. For this reason, pure image-based approaches have not been considered so far. We show that this challenge can be overcome by splitting the generating process in two parts. First, we learn to generate a semantic segmentation of the body and clothing. Second, we learn a conditional model on the resulting segments that creates realistic images. The full model is differentiable and can be conditioned on pose, shape or color. The result are samples of people in different clothing items and styles. The proposed model can generate entirely new people with realistic clothing. In several experiments we present encouraging results that suggest an entirely data-driven approach to people generation is possible.
\end{abstract}
\renewcommand*{\thefootnote}{\fnsymbol{footnote}}
\footnotetext{\textsuperscript{*} This work was performed while P. V. Gehler was with the BCCN\textsuperscript{1} and MPI-IS\textsuperscript{2}.}
\renewcommand*{\thefootnote}{\arabic{footnote}}
\setcounter{footnote}{0}

\section{Introduction}\label{sec:introduction}

% Motivation.
Perceiving people in images is a long standing goal in computer vision. Most work focuses on detection, pose and shape estimation of people from images. In this paper, we address the inverse problem of automatically generating images of people in clothing. A traditional approach to this task is to use computer graphics. A pipeline including 3D avatar generation, 2D pattern design, physical simulation to drape the cloth, and texture mapping is necessary to render an image from a 3D scene.

% Why not graphics?
Graphics pipelines provide precise control of the outcome. Unfortunately, the rendering process poses various challenges, all of which are active research topics and mostly require human input.
Especially clothing models require expert knowledge and are laborious to construct:
the physical parameters of the cloth must be known in order to achieve a realistic result. In addition, modeling the complex interactions between the body and clothing and between different layers of clothing presents challenges for many current systems. The overall cost and complexity limits the applications of realistic cloth simulation.
Data driven models of cloth can make the problem easier, but available data of clothed people in 3D is scarce.

% Our approach.
% Nomenclature introduction.
Here, we investigate a different approach and aim to sidestep this pipeline. We propose \cn/, a generative model of people learned directly from images. \cn/ uses task specific information in the form of a 3D body model, but is mostly data-driven. A basic version (\emph{\cnplain/}) allows to randomly generate images of people from a learned latent space. To provide more control we also introduce a conditional model (\emph{\cncond/}). Given a synthetic image silhouette of a projected 3D body model, \cncond/ produces random people with similar pose and shape in different clothing styles (see Fig.~\ref{fig:teaser}).

Learning a direct image based model has several advantages: firstly, we can leverage large photo-collections of people in clothing to learn the statistics of how clothing maps to the body; secondly, the model allows to dress people fully automatically, producing plausible results. Finally, the model learns to add realistic clothing accessories such as bags, sunglasses or scarfs based on image statistics.

We run multiple experiments to assess the performance of the proposed models. Since it is inherently hard to evaluate metrics on generative models, we show representative results throughout the paper and explore the encoded space in a principled way in several experiments. To provide an estimate on the perceived quality of the generated images, we conducted a user study. With a rate of 24.7\textperc or more, depending on the \cn/ variant, humans take the generated images for real.

%%% Local Variables:
%%% mode: latex
%%% TeX-master: "../paper"
%%% End:

%  LocalWords:  variational autoencoders

\vspace*{-0.2cm}
\section{Related Work}\label{sec:related_work}
\vspace*{-0.1cm}
\subsection{3D Models of People in Clothing}\label{sec:3d-models-of-people}
\vspace*{-0.1cm}
\begin{figure}
  \newlength{\figRelatedBorder}
  \setlength{\figRelatedBorder}{0.cm}
  \newlength{\figRelatedHeight}
  \setlength{\figRelatedHeight}{3cm}
  \begin{center}
    \includegraphics[height=\figRelatedHeight]{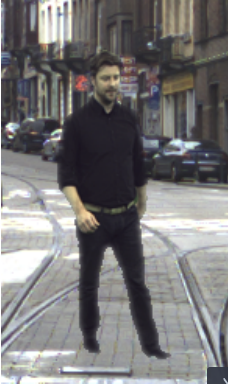}%
    \includegraphics[height=\figRelatedHeight]{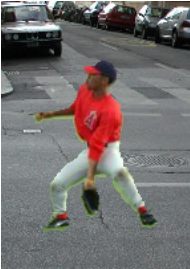}%
    \includegraphics[height=\figRelatedHeight]{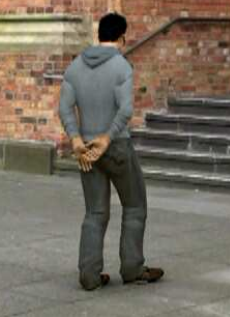}\\%
    \includegraphics[height=\figRelatedHeight]{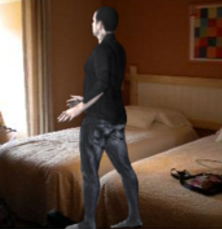}%
    \hspace*{-0.21cm}%
    \includegraphics[height=\figRelatedHeight]{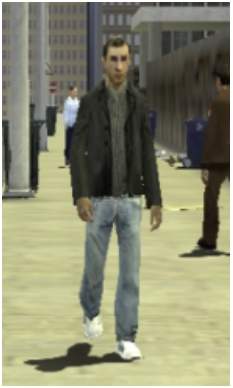}%
    \includegraphics[height=\figRelatedHeight]{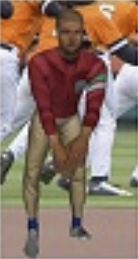}%
  \end{center}
  \vspace*{-0.5cm}
  \caption{Sample results for virtual humans from existing approaches (ltr., ttb.): \cite{leonid11cvpr}, \cite{pishchulin12reshaping}, local warping. \cite{h36m_pami}, \cite{learning_from_synthetic}, animated 3D scans in real environments. \cite{virtual_humans}, 3D avatars in virtual environments. \cite{synth3Dpose}, 3D avatars in real environments.
    % For a detailed discussion, see Sec.~\ref{sec:3d-models-of-people}.
    \label{fig:rel-results}
  }%
  \vspace*{-0.5cm}
\end{figure}

There exists a large and diverse literature on the topic of creating realistic looking images of people. They can be grouped into rendering systems and systems that attempt to modify existing real photographs (warping pixels).

\paragraph{Warping pixels.}
Xu et al.~\cite{Xu:2011:VCC} pose the problem as one of video retrieval and warping. Rather than synthesizing meshes with wrinkles, they look up video frames with the right motions. Similarly, in~\cite{Jain:2010,zhou2010parametric} an unclothed body model is fit to multi-camera and monocular image data. The body is warped and the image reshaped accordingly.

Two prominent works that aim to reshape people in photos are~\cite{pishchulin12reshaping,leonid11cvpr} (\cf Fig.~\ref{fig:rel-results}). A number of different synthetic sources has been used in~\cite{leonid11cvpr} for improvement of pedestrian detection. The best performing source is obtained morphing images of people, but requires data from a multi-view camera setup; consequently only 11 subjects were used. Subsequent work~\cite{pishchulin12reshaping} reshaped images but required significant user interaction. All aforementioned approaches require manual input and can only modify existing photographs.

\vspace*{-0.5cm}
\paragraph{Rendering systems.}
Early works synthesizing people from a body model are~\cite{PonsTaylorMetricForests2013,shotton2013real,taylor2012vitruvian}. Here, renderings were limited to depth images with the goal of improving human pose estimation. The work of~\cite{synth3Dpose} combines real photographs of clothing with a SCAPE~\cite{SCAPE} body model to generate synthetic people whose pose can be changed (\cf Fig.~\ref{fig:rel-results}). The work of~\cite{rogez16nips} proposes a pose-aware  blending of 2D images, limiting the ability to generalize.

A different line of works use rendering engines with different sources of inputs. In~\cite{h36m_pami}, a mixed reality scenario is created by rendering 3D rigged animation models into videos, but it is clearly visible that results are synthetic (\cf Fig.~\ref{fig:rel-results}). The work of~\cite{learning_from_synthetic} combines a physical rendering engine together with real captured textures to generate novel views of people. All these works use 3D body models without clothing geometry, hence the quality of the results is limited. One exception is~\cite{guan20102d} where only the 2D contour of the projected cloth is modeled.

\vspace*{-0.5cm}
\paragraph{3D clothing models.}
Much of the work in the field of clothing modeling is focused on how to make simulation faster~\cite{GHFBG07,Narain:2012:AAR}, particularly by adding realistic wrinkles to low-resolution simulations~\cite{Kavan:2011,Kim:2013:NEP}. Other approaches have focused on taking off-line simulations and learning data driven models from them~\cite{deAguiar:2010:SSR,DRAPE:2012,Kim:2013:NEP,Stoll:2010,Wang:SIGGRAPH:2011}. The authors of \cite{rogge2014} simulate clothing in 3D and project back to the image for augmentation. All these approaches require pre-designed garment models. Furthermore, current 3D models are not fully automatic, restricted to a set of garments or not photorealistic. The approach of~\cite{Pons-Moll:Siggraph2017} automatically captures real clothing, estimates body shape and pose~\cite{shape_under_cloth:CVPR17} and retargets to new body shapes. The approach does not require predefined 3D garments but requires a 3D scanner. 

\subsection{Generative Models}

\paragraph{Variational models and GANs.}
Variational methods are a well-principled approach to build generative models. Kingma and Welling developed the Variational Autoencoder~\cite{vae}, which is a key component of our method. In their original work, they experimented with a multi-layer perceptron on low resolution data. Since then, multiple projects have designed VAEs for higher resolutions, \eg,~\cite{attribute2image}. They use a CVAE~\cite{cvae,cvae2} to condition generated images on vector embeddings.

Recurrent formulations~\cite{draw,pixelrnn,pixcnn} enable to model complex structures, but again only at low resolution. With~\cite{lapgan}, Denton et al. address the resolution issue explicitly and propose a general architecture that can be used to improve network output. This strategy could be used to enhance \cn/. Generative Adversarial Networks~\cite{gans} use a second network during the training to distinguish between training data and generated data to enhance the loss of the trained network. We use this strategy in our training to increase the level of detail of the generated images. Most of the discussed works use resolutions up to 64x64 while we aim to generate 256x256 images. For our model design we took inspiration from encoder-decoder architectures such as the U-Net~\cite{unet}, Context Encoders~\cite{context_encoders} and the image-to-image translation networks~\cite{pix2pix}.

\vspace*{-.5cm}
\paragraph{Inpainting methods.}
Recent inpainting methods achieve a considerable level of detail~\cite{context_encoders,yang2016high} in resolution and texture. To present a comparison with a state-of-the art encoder-decoder architecture, we provide a comparison with~\cite{context_encoders} in our experiments. \cite{saito2016photorealistic} works directly on a texture map of a 3D model. Future work could explore to combine it with our approach from 2D image databases.

\vspace*{-.5cm}
\paragraph{Deep networks for learning about 3D objects.}
There are several approaches to reason about 3D object configuration with deep neural networks. The work of Kulkarni~\cite{kulkarni2015deep} use VAEs to model 3D objects with very limited resolution and assume that a graphics engine and object model are available at learning time. In~\cite{dai2016shape} an encoder-decoder CNN in voxel space is used for 3D shape completion, which requires 3D ground truth. The authors of~\cite{oberweger2015training} develop a generative model to create depth training data for articulated hands. This avoids the problem of generating realistic appearance.

%%% Local Variables:
%%% mode: latex
%%% TeX-master: "../paper"
%%% End:

\vspace*{-.2cm}
\section{Chictopia made SMPL}\label{sec:dataset}
\vspace*{-.2cm}

To train a supervised model connecting body parameters to fashion, we need a dataset providing information about both. Datasets for training pose estimation systems~\cite{mpiihp,h36m_pami,lsp} capture complex appearance, shape and pose variation, but are not labeled with clothing information. The Chictopia10K dataset~\cite{chictopia10k} contains fine-grained fashion labels but no human pose and shape annotations. In the following sections, we explain how we augmented Chictopia10K in an automatic manner so that it can be used as a resource for generative model training.

\begin{figure}
  \vspace*{-.1cm}
  \newlength{\figFitBorder}
  \setlength{\figFitBorder}{0.2cm}
  \newlength{\figFitHeight}
  \setlength{\figFitHeight}{3.5cm}
  \begin{center}
    \hspace*{\figFitBorder}
    \includegraphics[height=\figFitHeight]{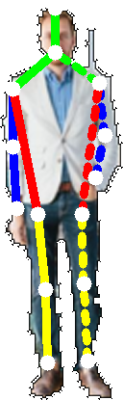}%
    \hspace*{-0.1cm}%
    \includegraphics[height=\figFitHeight]{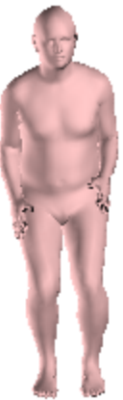}%
    \hfill%
    \includegraphics[height=\figFitHeight]{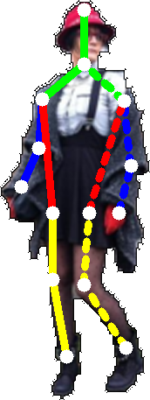}%
    \hspace*{-0.3cm}%
    \includegraphics[height=\figFitHeight]{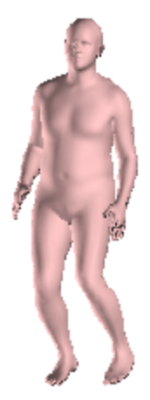}%
    \hfill%
    \includegraphics[height=\figFitHeight]{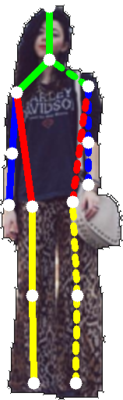}%
    \hspace*{-0.2cm}
    \includegraphics[height=\figFitHeight]{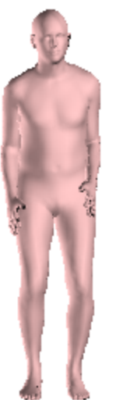}%
    \hspace*{\figFitBorder}%
  \end{center}
  \vspace*{-0.6cm}
  \caption{Example images from the Chictopia10K dataset~\cite{chictopia10k}, detected joints from DeeperCut~\cite{deepercut} and the final SMPLify fits~\cite{smplify}. Typical failure cases are foot and head orientation (\textbf{center}). The pose estimator works reliably even with wide clothing and accessories. (\textbf{right}).\label{fig:smpl_fit_examples}}
  \vspace*{-0.2cm}
\end{figure}

\begin{figure}
  \newlength{\figdsetenhBorder}
  \setlength{\figdsetenhBorder}{0.1cm}
  \newlength{\figdsetenhHeight}
  \setlength{\figdsetenhHeight}{3.2cm}
  \begin{center}
    \hspace*{\figdsetenhBorder}
    \includegraphics[height=\figdsetenhHeight]{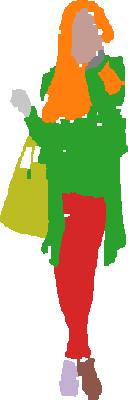}
    \hspace*{-0.2cm}
    \includegraphics[height=\figdsetenhHeight]{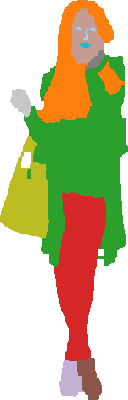}
    \hfill
    \includegraphics[height=\figdsetenhHeight]{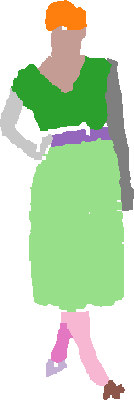}
    \hspace*{-0.2cm}
    \includegraphics[height=\figdsetenhHeight]{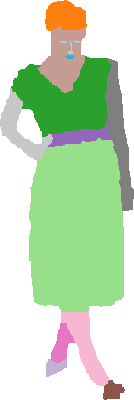}
    \hfill
    \includegraphics[height=\figdsetenhHeight]{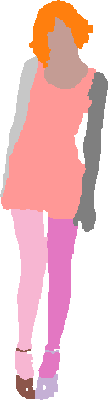}
    \hspace*{-0.2cm}
    \includegraphics[height=\figdsetenhHeight]{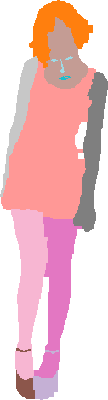}
    \hspace*{\figdsetenhBorder}
  \end{center}
  \vspace*{-0.6cm}
  \caption{Example annotations from the Chictopia10K dataset~\cite{chictopia10k} before and after processing (for each pair \textbf{left} and \textbf{right} respectively). Holes are inpainted and a face shape matcher is used to add facial features. The rightmost example shows a failure case of the face shape matcher.\label{fig:dataset_enhancement}}
  \vspace*{-0.5cm}
\end{figure}

\vspace*{-.4cm}
\paragraph{Fitting SMPL to Chictopia10K.}
The \chic~dataset consists of 17,706 images collected from the chictopia fashion blog\footnote{\url{http://www.chictopia.com/}}. For all images, a fine-grained segmentation into 18 different classes (\cf~\cite{chictopia10k}) is provided: 12 clothing categories, background and 5 features such as hair and skin, see Fig.~\ref{fig:dataset_enhancement}. For shoes, arms and legs, person centric left and right information is available. We augment the \chic~dataset with pose and shape information by fitting the 3D SMPL body model~\cite{smpl} to the images using the SMPLify pipeline~\cite{smplify}. SMPLify requires a set of 2D keypoint locations which are computed using the DeeperCut~\cite{deepercut} pose estimator.

Qualitative results of the fitting procedure are shown in Fig.~\ref{fig:smpl_fit_examples}. The pose estimator has a high performance across the dataset and the 3D fitting produces few mistakes. The most frequent failures are results with wrong head and foot orientation. To leverage as much data as possible to train our supervised models, we refrain from manually curating the results. Since we are interested in overall body shape and pose, we are using a six-part segmented projection of the SMPL fits for conditioning of \cncond/. Due to the rough segmentation, segmented areas are still representative even if orientation details do not match.

\vspace*{-0.4cm}
\paragraph{Face Shape Matching and Mask Improvement.}

We further enhance the annotation information of Chictopia10K and include facial landmarks to add additional guidance to the generative process. With only a single label for the entire face, we found that all models generate an almost blank skin area in the face.

We use the dlib~\cite{dlib} implementation of the fast facial shape matcher~\cite{face_alignment} to enhance the annotations with face information. We reduce the detection threshold to oversample and use the face with the highest intersection over union (IoU) score of the detection bounding box with ground truth face pixels. We only keep images where either no face pixels are present or the IoU score is above a certain threshold. A threshold score of 0.3 was sufficient to sort out most unusable fits and still retain a dataset of 14,411 images~(81,39\textperc).

Furthermore, we found spurious ``holes'' in the segmentation masks to be problematic for the training of generative models. Therefore, we apply the morphological ``close'' and ``blackhat'' operations to fill the erroneously placed background regions. We carefully selected the kernel size and found that a size of 7 pixels fixes most mistakes while retaining small structures. You can find examples of original and processed annotations in Fig.~\ref{fig:dataset_enhancement}.

%%% Local Variables:
%%% mode: latex
%%% TeX-master: "../paper"
%%% End:

%  LocalWords:  keypoints inpainted Chictopia matcher

\section{ClothNet}\label{sec:model}

\begin{figure*}
  \vspace*{-0.4cm}
  \newlength{\figModulesBorder}
  \setlength{\figModulesBorder}{0.2cm}
  \newlength{\figModulesBotBorder}  % Doesnt work at the moment.
  \setlength{\figModulesBotBorder}{-2cm}
  \hspace*{\figModulesBorder}
  \subfloat[]{
    \begin{minipage}[c]{0.28\textwidth}
      \centering
      \vspace*{-4.3cm}
      \includegraphics[width=\textwidth]{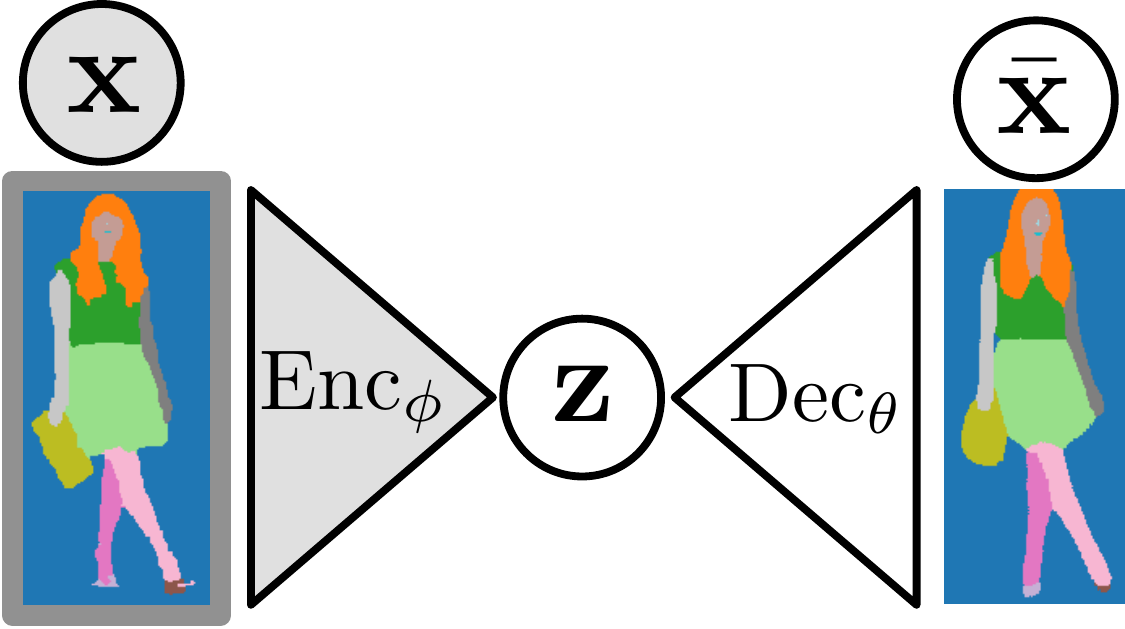}
    \end{minipage}
    \vspace*{\figModulesBotBorder}
  }
  \hfill
  \subfloat[]{
    \includegraphics[width=0.28\textwidth]{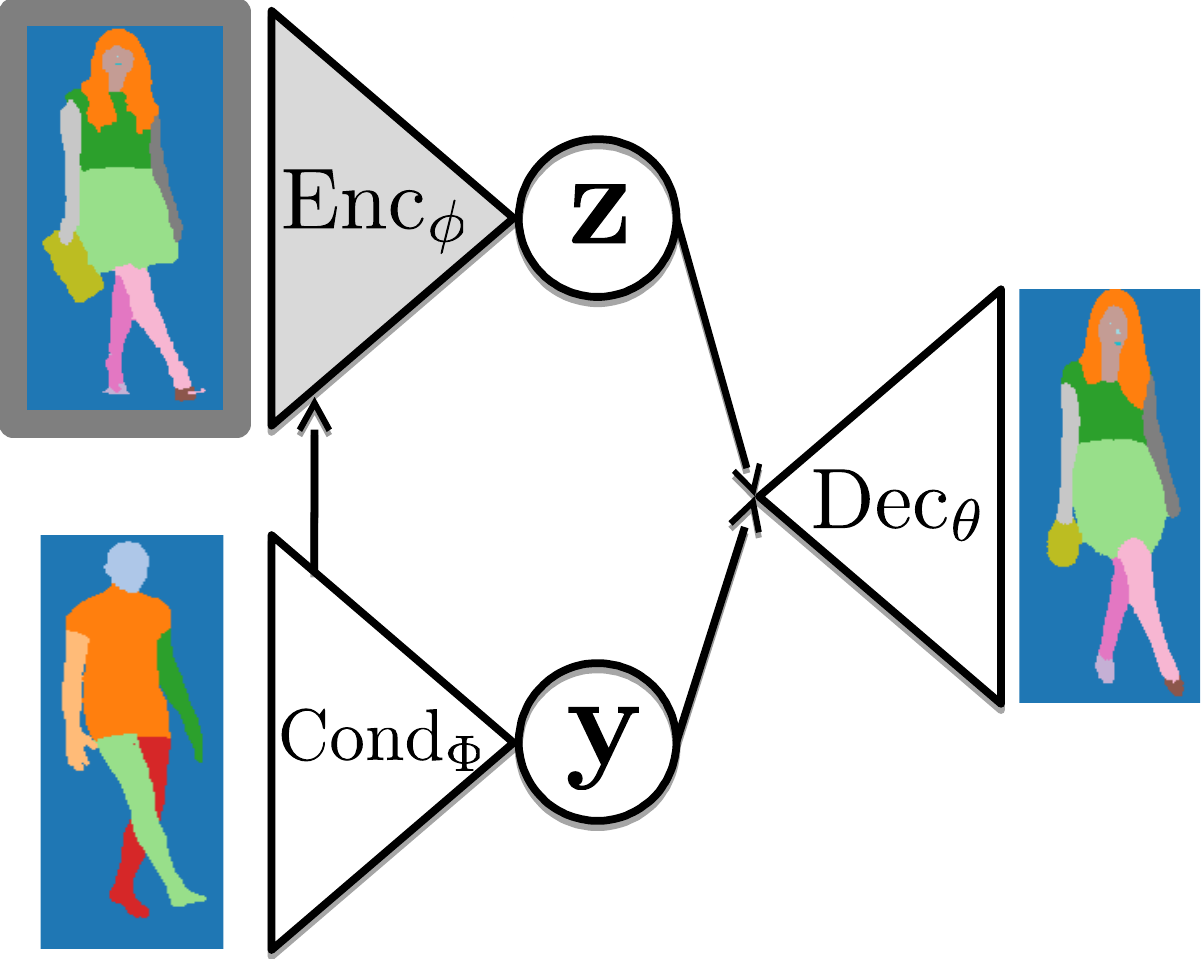}
    \vspace*{\figModulesBotBorder}
  }
  \hfill
  \subfloat[]{
    \begin{minipage}[c]{0.28\textwidth}
      \centering
      \vspace*{-4.3cm}
      \includegraphics[width=0.95\textwidth]{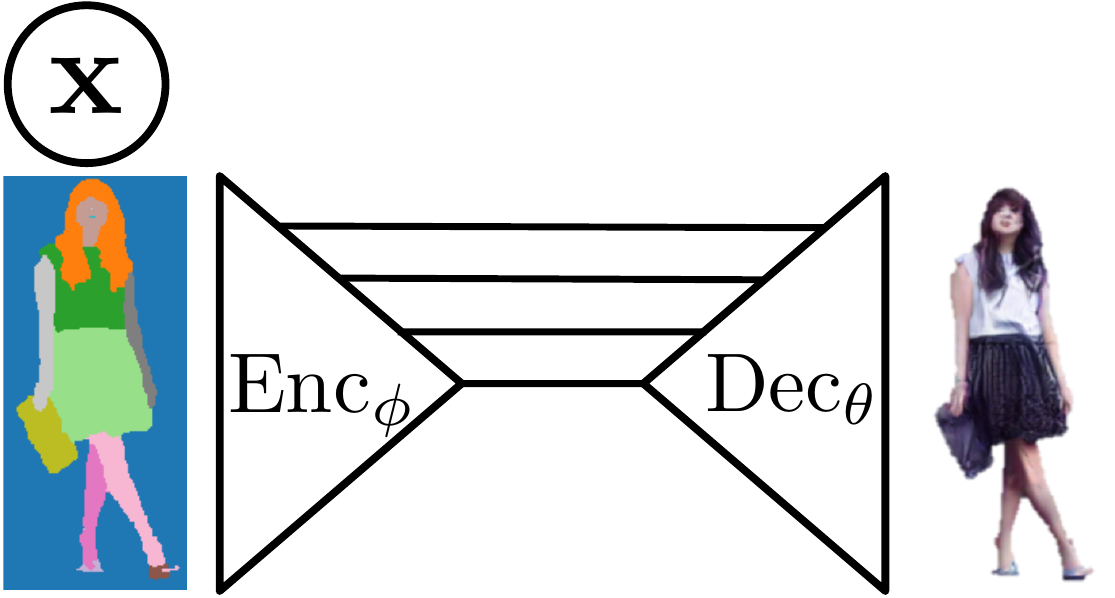}
    \end{minipage}
    \vspace*{\figModulesBotBorder}
  }
  \hspace*{\figModulesBorder}
  \vspace*{-0.2cm}
  \caption{\cn/ modules: \textbf{(a)} the latent sketch module consists of a variational auto-encoder, \textbf{(b)} the conditional sketch module
  consists of a conditional variational auto-encoder and \textbf{(c)} the \portray/ module is an image-to-image translation network that fills a sketch with texture.
  The modules in (a) and (c) are concatenated to form \cnplain/ and modules (b) and (c) are concatenated to form \cncond/.
  The learned latent representation $\mathbf{z}$ in (a) and (b) is a 512-D random variable that follows a multivariate Gaussian distribution. The variable
  $\mathbf{y}$ is a deterministic latent encoding of the body model silhouette that we use to condition on pose and shape.
  At test time in (a) and (b) one can generate a sample from the multivariate Gaussian $\mathbf{z}^i \sim \mathcal{N}(\mathbf{0}, \mathbf{I})$ and push them through the decoder network to produce
  random sketches of people in different clothing. We show in (a) and (b) the input to the encoder in gray color, indicating that they are not available at test time.
  \label{fig:modules}}
\vspace*{-0.4cm}
\end{figure*}

Currently, the most visually appealing technique to create fine-grained textures at $256\times256$ resolution are image-to-image translation networks~\cite{pix2pix}. An encoder-decoder structure with skip connections between their respective layers allows the model to retain sharp edges.

\begin{figure}
  \newlength{\figBaselineBorder}
  \setlength{\figBaselineBorder}{0.2cm}
  \newlength{\figBaselineHeight}
  \setlength{\figBaselineHeight}{3.2cm}
  \begin{center}
    \hspace*{\figBaselineBorder}
    \subfloat[]{
      \includegraphics[height=\figBaselineHeight]{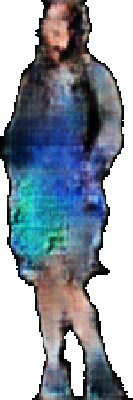}
      \includegraphics[height=\figBaselineHeight]{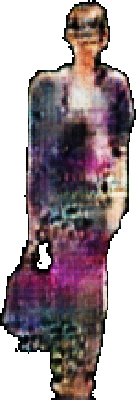}
      \includegraphics[height=\figBaselineHeight]{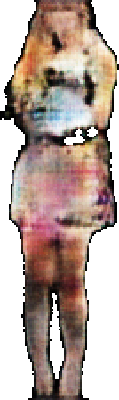}
    }
    \hfill
    \subfloat[\label{fig:results_pix2pix_direct}]{
      \includegraphics[height=\figBaselineHeight]{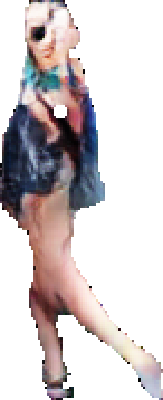}
      \includegraphics[height=\figBaselineHeight]{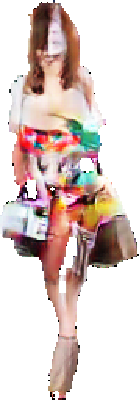}
      \includegraphics[height=\figBaselineHeight]{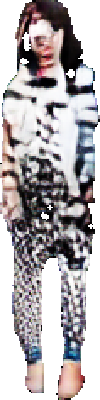}
    }
    \hspace*{\figBaselineBorder}
  \end{center}
  \vspace*{-0.5cm}
  \caption{Example results from \textbf{(a)} the context encoder architecture~\cite{context_encoders} from a ground truth sketch. Without skip connections, the level of predicted detail remains low. \textbf{(b)} Results from an image-to-image network trained to predict dressed people from six part SMPL sketches directly. Without the proposed two-stage architecture, the model is not able to determine shape and cloth boundaries.\label{fig:results_pix2pix_ce}}
  \vspace*{-0.5cm}
\end{figure}

However, applying image-to-image translation networks directly to the task of predicting dressed people from SMPL sketches as displayed in Fig.~\ref{fig:teaser} does not produce good results (we provide example results from such a model in Fig.~\ref{fig:results_pix2pix_direct}). The reason is that there are many possible completions for a single 3D pose. The image-to-image translation model can not handle such multi-modality. Furthermore, its sampling capabilities are poor. Variational Autoencoders, on the other hand, excel at encoding high variance for similar inputs and provide a principled way of sampling.

We combine the strengths of both, Variational Autoencoders and the image-to-image translation models, by stacking them in a two-part model: the \textit{sketch} part is variational and deals with the high variation in clothing shape. Its output is a semantic segmentation map (sketch) of a dressed person. The second \textit{\portray/} part uses the created sketch to generate an image of the person and can make of use of skip connections to produce a detailed output. In the following sections, we will introduce the modules we experimented with.

\subsection{The Latent Sketch Module}\label{sec:sketch_module}
The latent sketch module is a variational auto-encoder which allows to sample random sketches of people.

\paragraph{The Variational Auto-Encoder~\cite{vae}} consists of two parts, an \textit{encoder} to a latent space, and a \textit{decoder} from the latent space to the original representation. As for any latent variable model, the aim is to reconstruct the training set $\mathbf{x}$ from a latent representation $\mathbf{z}$. Mathematically, this means maximizing the data likelihood $p(\mathbf{x}) = \int p_\theta(\mathbf{x}|\mathbf{z})p(\mathbf{z})d\mathbf{z}$. In high dimensional spaces, finding the decoder parameters $\theta$ that maximize the likelihood is intractable. However, for many values of $\mathbf{z}$ the probability $p_\theta(\mathbf{x}|\mathbf{z})$ will be almost zero. This can be exploited by finding a function $q_\phi(\mathbf{z}|\mathbf{x})$, the \emph{encoder}, parameterized by $\phi$. It encodes a sample $\mathbf{x}^i$ and produces a distribution over $z$ values that are likely to reproduce $\mathbf{x}^i$. To make the problem tractable and differentiable, this distribution is assumed to be Gaussian, $q_\phi(\mathbf{z}|\mathbf{x}^i) = \mathcal{N}(\bm{\mu}^i, \bm{\Sigma}^i)$. The parameters $\bm{\mu}^i, \bm{\Sigma}^i$ are predicted by the $\phi$-parameterized encoding neural network $\text{Enc}_{\phi}$. The decoder is the $\theta$ parameterized neural network $\text{Dec}_{\theta}$.

Another key assumption for VAEs is that the marginal distribution on the latent space is Gaussian distributed with zero mean and identity covariance, $p(\mathbf{z}) = \mathcal{N}(\mathbf{0},\mathbf{I})$. Under these assumptions, the VAE objective (see \cite{vae} for derivations) to be maximized is
\begin{equation}
\sum_i E_{z \sim q}[\log p_\theta(\mathbf{x}^i|\mathbf{z})] - D_{KL}(q_\phi(\mathbf{z}|\mathbf{x}^i) || p(\mathbf{z}))
\label{eq:VAE_objective},
\end{equation}
where $E_{z \sim q}$ indicates expectation over distribution $q$ and $D_{KL}$ denotes Kullback-Leibler (KL) divergence. The first term measures the decoder accuracy for the distribution produced by the encoder $q_\phi (\mathbf{z}|\mathbf{x})$ and the second term penalizes deviations of $q_\phi(\mathbf{z}|\mathbf{x}^i)$ from the desired marginal distribution $p(\mathbf{z})$. Intuitively, the second term prevents the encoding from carrying too much information about the input $\mathbf{x}^i$. Since both $q_\phi(\mathbf{z}|\mathbf{x}^i)$ and $p(\mathbf{z})$ are Gaussian, the KL divergence can be computed in closed form~\cite{vae}. Eq.~\eqref{eq:VAE_objective} is maximized using stochastic gradient ascent.

Computing Eq.~\eqref{eq:VAE_objective} involves sampling; constructing a sampling layer in the network would result in a non differentiable operation. This can be circumvented using the reparameterization trick~\cite{vae}. With this adaptation, the model is deterministic and differentiable with respect to the network parameters $\theta, \phi$. The latent space distribution is forced to follow a Gaussian distribution $\mathcal{N}(\mathbf{0},\mathbf{I})$ during training. This implies that at test time one can easily generate samples $\bar{\mathbf{x}}^i$ by generating a latent sample $\mathbf{z}^i \sim \mathcal{N}(\mathbf{0},\mathbf{I})$ and pushing it through the decoder $\bar{\mathbf{x}}^i=\text{Dec}_\theta(\mathbf{z}^i)$. This effectively ignores the encoder at test time.

\vspace*{-0.1cm}
\paragraph{Sketch encoding:} we want to encode images $\mathbf{x} \in \mathbb{R}^{256\times{256}}$ of sketches of people into a 512-D latent space, $\mathbf{z} \in \mathbb{R}^{512}$. This resolution requires a sophisticated encoder and decoder layout. Hence, we combine the recently proposed encoder and decoder architecture for image-to-image translation networks~\cite{pix2pix} with the formulation of the VAE. We use a generalized Bernoulli distribution to model $p_\theta(\mathbf{x}^i|\mathbf{z}).$ The architecture is illustrated in Fig.~\ref{fig:modules}(a).

\subsection{The Conditional Sketch Module\label{sec:cond-pose-shape}}

For some applications it may be desirable to generate different people in different clothing in a pose and shape specified by the user. To that end, we propose a module that we call \textit{conditional sketch module}.

The conditional sketch module gives control of pose and shape by conditioning on a 3D body model sketch as illustrated in Fig.~\ref{fig:modules}(b). We use a conditional variational autoencoder for this model (for a full derivation and description of the idea, we refer to~\cite{cvae}). To condition on an image $\mathbf{Y} \in \mathbb{R}^{256\times{256}}$ (a six part body model silhouette), the model is extended with a new encoding network $\text{Cond}_\Phi$, with similar structure as $\text{Enc}_\phi$. Since the conditioning variable is deterministic, the encoding is $\mathbf{y}= \text{Cond}_\Phi(\mathbf{Y})$. To provide the conditioning input to the encoder, we concatenate the output of the first layer of $\text{Cond}_\Phi$ to the output of the first layer of $\text{Enc}_\phi$. To train the model, we use the same objective as in Eq.~\eqref{eq:VAE_objective}. Here, the decoder reconstructs a sample using both, $\mathbf{z}$ and $\mathbf{y}$, with $\bar{\mathbf{x}}^i=\text{Dec}_\theta(\mathbf{y}^i,\mathbf{z}^i)$ and the minimization of the KL-divergence term is only applied to $\mathbf{z}$.

\subsection{The \portray/ Module\label{sec:portray-module}}

For applications requiring a textured image of a person, the sketch modules can be chained to a \portray/ module. We use an image-to-image translation model~\cite{pix2pix} to color the results from the sketch modules. With additional face information, we found this model to produce appealing results.

\subsection{\cnplain/ and \cncond/}\label{sec:ClothNet}

Once the sketch part and the \portray/ part are trained, they can be concatenated to obtain a full generative model of images of dressed people. We refer to the concatenation of the latent sketch module with the \portray/ module as \cnplain/. The concatenation of the conditional sketch module with the \portray/ module is named \cncond/. Several results produced by \cncond/ are illustrated in Fig.~\ref{fig:teaser}. All stages of \cnplain/ and \cncond/ are differentiable and implemented in the same framework. We trained the sketch and \portray/ modules separately, simply because it is technically easier; propagating gradients through the full model is possible\footnote{To propagate gradients through the full model, it must represent sketches as $256\times256\times22$ probability maps instead of $256\times256\times3$ color maps since applying the color map function is not differentiable in general. The \textit{portray} module results presented in the following sections have been created with color maps as inputs. The published code contains \textit{portray} models for both, color map and probability map inputs.}.

\vspace*{-.05cm}
\subsection{Network Architectures}\label{sec:impl-details}
\vspace*{-0.1cm}

Adhering to the image-to-image translation network architecture for designing encoders and decoders, we make use of LReLUs~\cite{lrelu}, batch normalization~\cite{batchnorm} and use fractionally strided convolutions~\cite{deconv}. We introduce weight parameters for the two loss components in Eq.~\eqref{eq:VAE_objective} and balance the losses by weighing the KL component with factor $6.55$. Then, the KL objective is optimized sufficiently well to create realistic samples $\bm{z}^i$ from $\mathcal{N}(\mathbf{0},\mathbf{I})$ after training. Full network descriptions are part of the supplementary material\footnote{\url{http://files.is.tue.mpg.de/classner/gp}}.

%%% Local Variables:
%%% mode: latex
%%% TeX-master: "../paper"
%%% End:

%  LocalWords:  parameterized variational segmentations autoencoder bayesian untractable differentiable

\section{Experiments}\label{sec:experiments}

\subsection{The Latent Sketch Module}
\vspace*{-0.2cm}
\begin{figure*}
  \newlength{\figLatentBorder}
  \setlength{\figLatentBorder}{0.2cm}
  \newlength{\figLatentHeight}
  \setlength{\figLatentHeight}{3.cm}
  \begin{center}
    \hspace*{\figLatentBorder}%
    \includegraphics[height=\figLatentHeight]{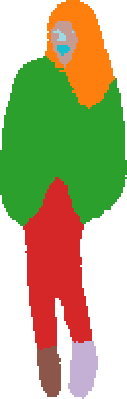}%
    \hfill%
    \includegraphics[height=\figLatentHeight]{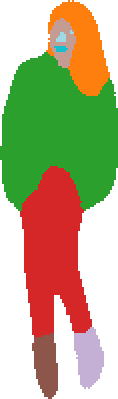}%
    \hfill%
    \includegraphics[height=\figLatentHeight]{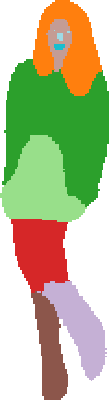}%
    \hfill%
    \includegraphics[height=\figLatentHeight]{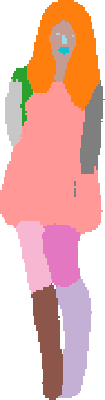}%
    \hfill%
    \includegraphics[height=\figLatentHeight]{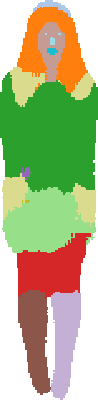}%
    \hfill%
    \includegraphics[height=\figLatentHeight]{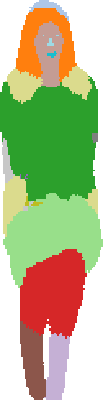}%
    \hfill%
    \includegraphics[height=\figLatentHeight]{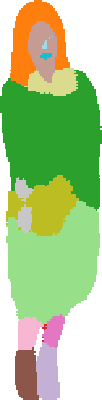}%
    \hfill%
    \includegraphics[height=\figLatentHeight]{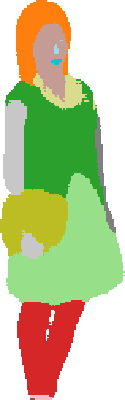}%
    \hfill%
    \includegraphics[height=\figLatentHeight]{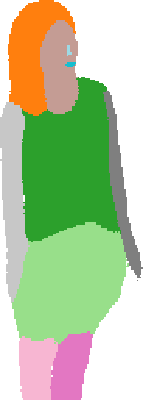}%
    \hfill%
    \includegraphics[height=\figLatentHeight]{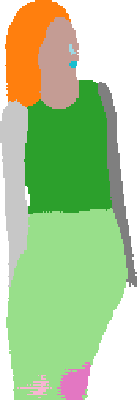}%
    \hspace*{\figLatentBorder}
  \end{center}
  \vspace*{-0.4cm}
  \caption{A walk in latent space along the dimension with the highest variance. We built a PCA space on the 512 dimensional latent vector predictions of the test set and walk -1STD to 1STD in equidistant steps.\label{fig:random_walk}}
  \vspace*{-0.2cm}
\end{figure*}

\begin{table}
  \vspace*{0.1cm}
  \begin{center}
    \resizebox{0.46\textwidth}{!}{%
      \begin{tabular}{|c|c|c|c|c|c|}
        \hline
        Model & Part & Accuracy & Precision & Recall & F1 \\
        \hline
        \hline
        \multirow{2}{*}{LSM}  & Train & 0.958 & 0.589 & 0.584 & 0.576 \\
        \cline{2-6}
                              & Test  & 0.952 & 0.540 & 0.559 & 0.510 \\
        \hline
        \multirow{2}{*}{CSM} & Train & 0.962 & 0.593 & 0.591 & 0.587 \\
        \cline{2-6}
                              & Test  & 0.950 & 0.501 & 0.502 & 0.488 \\
        \hline
      \end{tabular}%
    }
  \end{center}
  \vspace*{-0.5cm}
  \caption{Reconstruction metrics for the \textit{Latent Sketch Module} (CSM) and \textit{Conditional Sketch Module} (CSM). The overall reconstruction accuracy is high. The other metrics are dominated by classes with few labels. The CSM overfits faster.\label{tab:reconstruction_metrics}}
  \vspace*{-0.2cm}
\end{table}

Variational Autoencoders are usually evaluated on the likelihood bounds on test data. Since we introduced weights into our loss function as described in Sec.~\ref{sec:impl-details}, these would not be meaningful. However, for our purpose, the reconstruction ability of the sketch modules is just as important.

We provide numbers on the quality of reconstruction in Tab.~\ref{tab:reconstruction_metrics}. The values are averages of the respective metrics over all classes. The overall reconstruction accuracy is high with a score of more than 0.95 in all settings. The other metrics are influenced by the small parts, in particular facial features. The CSM overfits faster than the LSM due to the additional information from the conditioning.

For a generative model, qualitative assessment is important as well. For this, we provide a visualization of a high variance dimension in latent space in Fig.~\ref{fig:random_walk}. To create it, we produced the latent encodings $\mathbf{z}^i$ of all test set images. To normalize their distribution, we use the cumulative distribution function (CDF) values at their positions instead of the plain values. We then used a principal component analysis (PCA) to identify the direction with the most variance. In the PCA space, we take evenly spaced steps from minus one to plus one standard deviations; the PCA mean image is in the center of Fig.~\ref{fig:random_walk}.
Even though the direction encoding the most variance in PCA space only encodes roughly 1\textperc{} of the full variance, the complexity of the task becomes obvious: this dimension encodes variations in pose, shape, position, scale and clothing types. The model learns to adjust the face direction in plausible ways.

\subsection{The Conditional Sketch Module}\label{sec:cond-body}

As described in Sec.~\ref{sec:cond-pose-shape}, we use a CVAE architecture to condition the generated clothing segmentations. We use the SMPL body model to represent the conditioning. However, instead of using the internal SMPL vector representation of shape and pose, we render the SMPL body in the desired configuration. We use six body parts: head, central body, left and right arms, left and right legs, to give the model local cues about the body parts.

\begin{figure}
  \newlength{\figcondBorder}
  \setlength{\figcondBorder}{0.2cm}
  \newlength{\figcondHeight}
  \setlength{\figcondHeight}{3cm}
  \vspace*{-.7cm}
  \begin{center}
    \hspace*{\figcondBorder}
    \subfloat[]{\parbox{1.8cm}{%
        \begin{center}%
        \includegraphics[height=\figcondHeight]{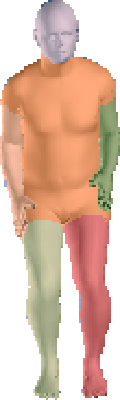}\\
        \includegraphics[height=\figcondHeight]{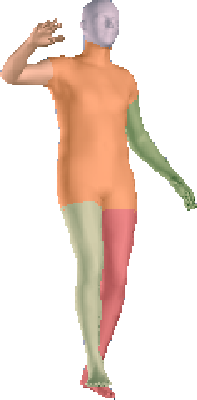}%
      \end{center}%
      \vspace*{-0.3cm}%
      }}
    \hfill
    \subfloat[]{\parbox{4.5cm}{%
        \begin{center}%
        \includegraphics[height=\figcondHeight]{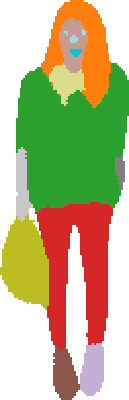}%
        \hspace*{0.08cm}%
        \includegraphics[height=\figcondHeight]{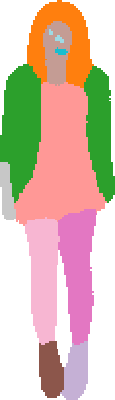}%
        \hspace*{0.08cm}%
        \includegraphics[height=\figcondHeight]{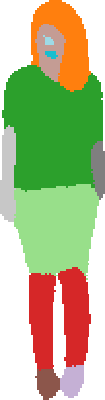}%
        \hspace*{0.08cm}%
        \includegraphics[height=\figcondHeight]{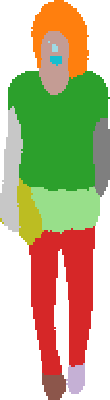}\\
        \includegraphics[height=\figcondHeight]{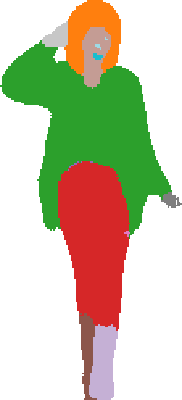}%
        \hspace*{-0.17cm}%
        \includegraphics[height=\figcondHeight]{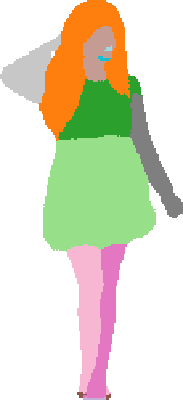}%
        \hspace*{-0.17cm}%
        \includegraphics[height=\figcondHeight]{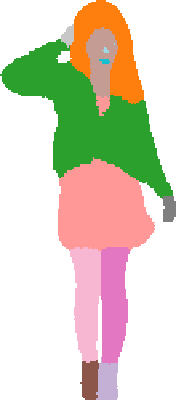}%
      \end{center}%
      \vspace*{-0.3cm}%
      }}
    \hfill
    \subfloat[]{\parbox{1.5cm}{%
        \begin{center}%
        \includegraphics[height=\figcondHeight]{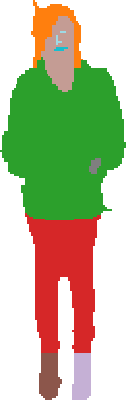}\\
        \includegraphics[height=\figcondHeight]{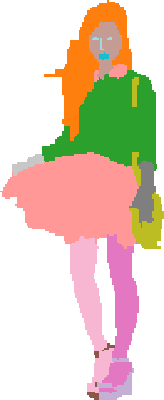}%
      \end{center}%
      \vspace*{-0.3cm}%
      }}
    \hspace*{\figcondBorder}
  \end{center}
  \vspace*{-0.5cm}
  \caption{Per row: \textbf{(a)} SMPL conditioning for pose and shape, \textbf{(b)} sampled dressed sketches conditioned on the same sample in (a), \textbf{(c)} the nearest neighbor of the rightmost sample in (b) from the training set. The model learns to add various hair types, style and accessories.}
  \label{fig:example_completions}
  \label{fig:pose_cond}
  \vspace*{-0.2cm}
\end{figure}

We found the six part representation to be a good trade-off: using only a foreground-background encoding may convey too little information, especially about left and right parts. A too detailed segmentation introduces too much noise, since the data for training our supervised models has been acquired by automatic fits solely to keypoints. These fits may not represent detailed matches in all cases. You can find qualitative examples of conditional sampling in Fig.~\ref{fig:teaser} and Fig.~\ref{fig:example_completions}.

At test time, we encode the model sketch (Fig.~\ref{fig:pose_cond}(a)) to obtain $\mathbf{y}^i = \text{Cond}_\Phi(\mathbf{Y})$, sample from the latent space $\mathbf{z}^i \sim p(\mathbf{z}) = \mathcal{N}(\mathbf{0},\mathbf{I})$ and obtain a clothed sketch with $\bar{\mathbf{x}}^i = \text{Dec}_\theta(\mathbf{y}^i,\mathbf{z}^i)$. For every sample $\mathbf{z}^i$ a new sketch $\bar{\mathbf{x}}^i$ is generated with different clothing but roughly the same pose and shape. Notice how different samples produce different hair and cloth styles as well as different configurations of accessories such as bags.

\subsection{Conditioning on Color\label{sec:conditioning-color}}

\begin{figure}
    \input{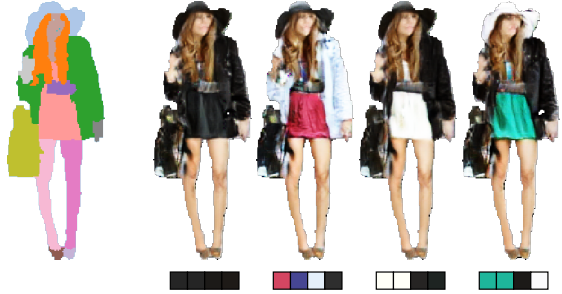}%
    \caption{Conditioning on color: (\textbf{left}) sketch input to the network. (\textbf{right}) Four different outputs for four different color combinations. Color conditioning for the regions are shown in the boxes below the samples (\textbf{boxes below, ltr}): lower dress, upper dress, jacket, hat.\label{fig:color_cond}}
    \vspace*{-0.5cm}
\end{figure}

As another example, we describe how to condition our model on color. During training, we compute the median color in the original image for every segment of a sketch. We create a new image by coloring the sketch parts with the respective median color. The concatenation of the colored image and the sketch are the new input to our \portray/ module which is retrained on this input. Conditioning can then be achieved by selecting a color for a sketch segment. An example result is shown in Fig.~\ref{fig:color_cond}. The network learns to follow the color cues, but still does not only generate plain color clothing, but places patterns, texture and wrinkles.

\subsection{\cn/}

With the following two experiments, we want to provide an insight in how realistic the images are that are generated from the \cnplain/ pipeline.

\vspace*{-0.3cm}
\subsubsection{Generating an Artificial Dataset}

\begin{figure*}
  \newlength{\figSynthDsetBorder}
  \setlength{\figSynthDsetBorder}{0.cm}
  \newlength{\figSynthDsetHeight}
  \setlength{\figSynthDsetHeight}{3.4cm}
  \begin{center}
    \hspace*{\figSynthDsetBorder}%
    \includegraphics[height=\figSynthDsetHeight]{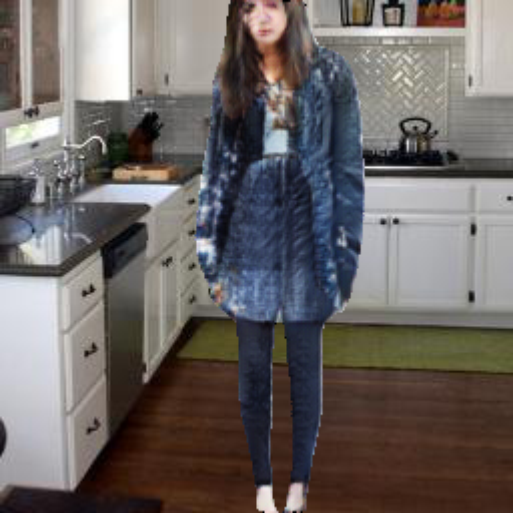}%
    \includegraphics[height=\figSynthDsetHeight]{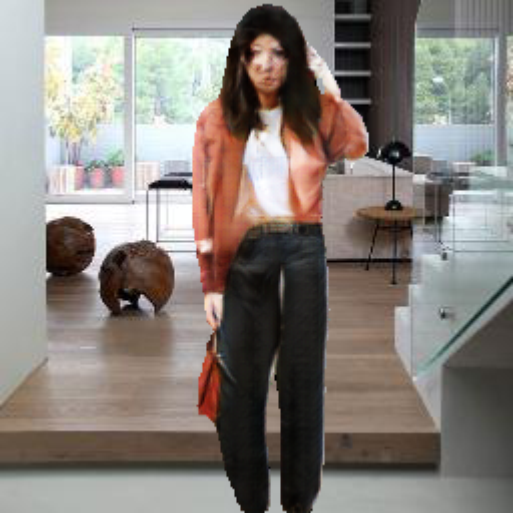}%
    \includegraphics[height=\figSynthDsetHeight]{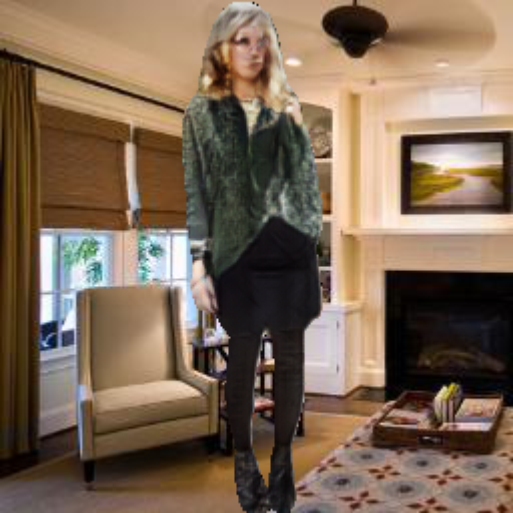}%
    \includegraphics[height=\figSynthDsetHeight]{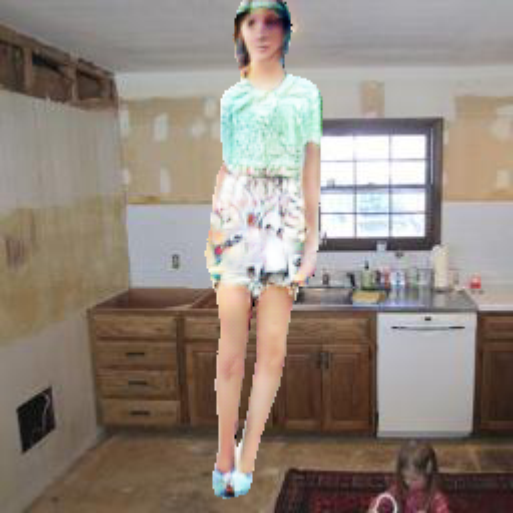}%
    \includegraphics[height=\figSynthDsetHeight]{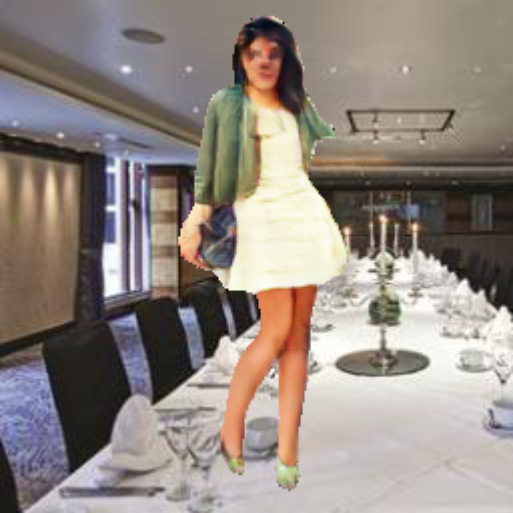}%
    \hspace*{\figSynthDsetBorder}\\
    \hspace*{\figSynthDsetBorder}%
    \includegraphics[height=\figSynthDsetHeight]{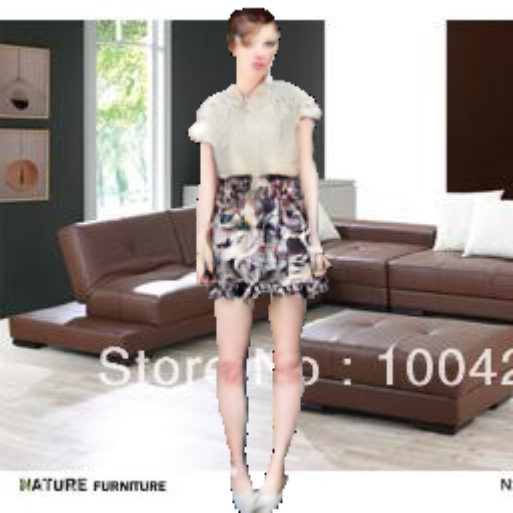}%
    \includegraphics[height=\figSynthDsetHeight]{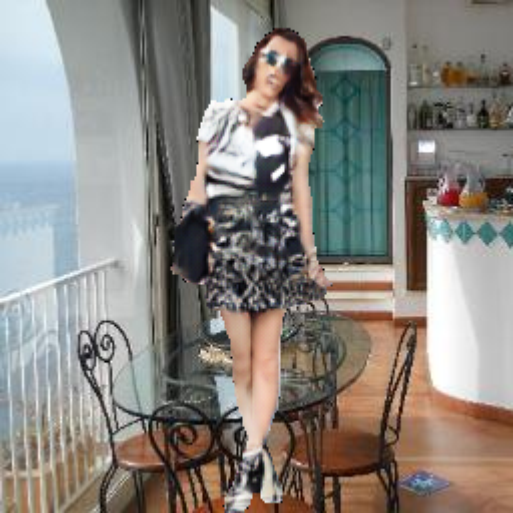}%
    \includegraphics[height=\figSynthDsetHeight]{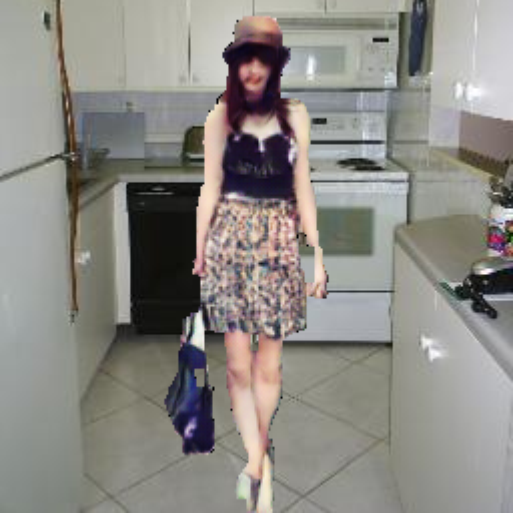}%
    \includegraphics[height=\figSynthDsetHeight]{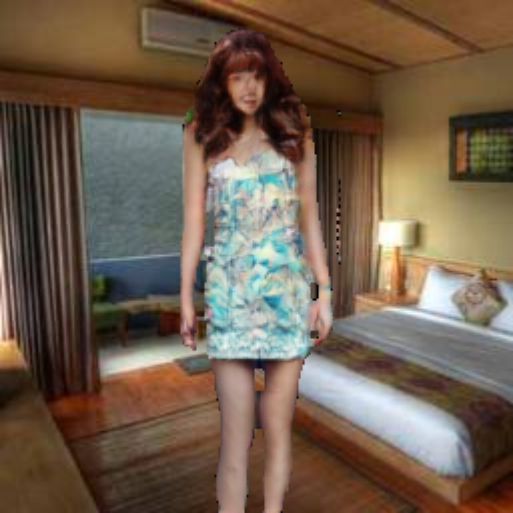}%
    \includegraphics[height=\figSynthDsetHeight]{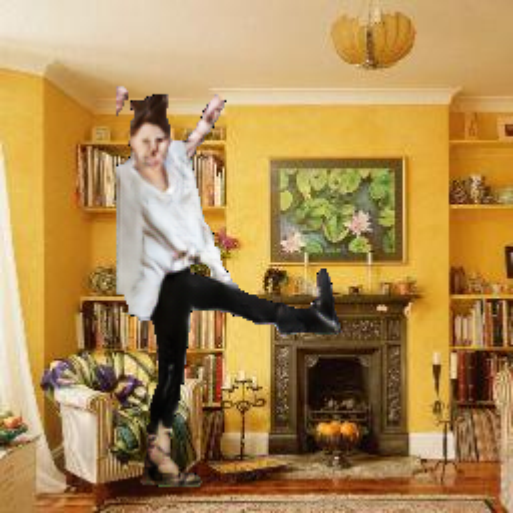}%
    \hspace*{\figSynthDsetBorder}%
  \end{center}
  \vspace*{-0.5cm}
  \caption{Results from \cn/ with added random backgrounds. \textbf{First row}: results from \cnplain/ (\ie, sketch and texture generation), \textbf{second row}: results from the \portray/ module on ground truth sketches.\label{fig:generative_results}}
  \vspace*{-0.2cm}
\end{figure*}

In the first experiment, we generate an artificial dataset and train a semantic segmentation network on the generated data. By comparing the performance of a discriminative model trained on real or synthetic images we can asses how realistic the generated images are.

For this purpose, we generate an equally sized dataset to our enhanced subset of Chictopia10K. We store the semantic segmentation masks generated from the latent sketch module as artificial `ground truth' and the outputs from the full \cnplain/ pipeline as images. To make the images comparable to the Chictopia10K images, we add artificial background. Similar to~\cite{learning_from_synthetic}, we sample images from the \textit{dining room}, \textit{bedroom}, \textit{bathroom} and \textit{kitchen} categories of the LSUN dataset~\cite{lsun}. Example images are shown in Fig.~\ref{fig:generative_results}.

Even though the generated segmentations from our VAE model look realistic at first glance, some weaknesses become apparent when completed by the \portray/ module: bulky arms and legs and overly smooth outlines of fine structures such as hair. Furthermore, the different statistics of facial landmark size to ground truth sketches lead to less realistic faces.%\footnote{We plan to address this issue by training a \portray/ module on the outputs of the sketch modules of the ground truth sketches.\pg{no idea} Results from this model could not be shown due to time reasons in this version of the paper.\pg{So maybe: In the future we plan to ... (simpler please)}}

\begin{table}
  \newcommand{\bc}[2][c]{%
    \begin{tabular}[#1]{@{}c@{}}#2\end{tabular}}%
  \begin{center}
    \begin{tabular}{|c|c|c|c|c|c|}
      \hline
      \backslashbox{Train}{Test} & Full Synth.         & Synth. Text.        &  Real \\
      \hline
      Full Synth.                & \bc{0.566\\0.978}   & \bc{0.437\\0.964}   & \bc{0.335\\0.898} \\
      \hline
      Synth. Text.               & \bc{0.503\\0.968}   & \bc{0.535\\0.976}   & \bc{0.411\\0.915} \\
      \hline
      Real                       & \bc{0.448\\0.955}   & \bc{0.417\\0.957}   & \bc{0.522\\0.951} \\
      %RL  & \bc{0.4333\\0.9522} & \bc{0.3990\\0.9549} & \bc{0.5056\\0.9486} \\
      \hline
    \end{tabular}
  \end{center}
  \vspace*{-0.5cm}
  \caption{Segmentation results (per line: intersection over union (IoU), accuracy) for a variety of training and testing datasets. \textbf{Full Synth.} results are from the \cnplain/ model, \textbf{Synth. Text.} from the \portray/ module on ground truth sketches.\label{tab:segmentation_results}}
\end{table}

We train a DeepLab ResNet-101 \cite{deeplab} segmentation model on real and synthetic data and evaluate on test images from all data sources. Evaluation results for this model can be found in Tab.~\ref{tab:segmentation_results}. As expected, the models trained and tested from the same data source perform best. The model trained on the real dataset reaches the highest performance and can be trained longest without overfitting. The fully synthetic datasets lose at most 5.3 accuracy points compared to the model trained on real data. The IoU scores, however, suffer from fewer fine structures present in the generated data such as sunglasses and belts.

\subsubsection{User Study}

\begin{table}
  \begin{center}
    \resizebox{0.46\textwidth}{!}{%
      \begin{tabular}{|c|c|c|}
        \hline
        Model & Real image rated gen. & Gen. image rated real \\
        \hline
        \hline
        \cnplain/ & 0.154 & 0.247 \\
        \hline
        \portray/ mod.  & 0.221 & 0.413 \\
        \hline
      \end{tabular}%
    }
  \end{center}
  \vspace*{-0.5cm}
  \caption{User study results from 12 participants. The first row shows results for the full \cnplain/ model, the second for the \portray/ module used on ground truth sketches.\label{tab:user_study}}
  \vspace*{-0.5cm}
\end{table}

% Problem.
We performed a user study to quantify the realism of images. We set up an experiment to evaluate both stages of our model: one for images generated from the \portray/ module on ground truth sketches and once for the full \cnplain/ model. For each of the experiments, we asked users to label 150 images for being a photo or generated from our model. 75 images were real Chictopia images, 75 generated with our model. Every image was presented for 1 second akin to the user study in Isola et al.~\cite{pix2pix}. We blanked out the faces of all images since those would be dominating the decision of the participants: this body part still provides the most reliable cues for artificially generated images. The first 10 rated images are ignored to let users calibrate on image quality.

With this setup we follow the setup of Isola et al.~\cite{pix2pix}. They use a forced choice between two images, one ground truth, one sketched by their model on ground truth segmentation. Since we do not have ground truth comparison images, we display one image at a time and ask for a choice. This setting is slightly harder for our model, since the user can focus on one image. The results for the 12 participants of our study are presented in Tab.~\ref{tab:user_study}. Even by the fully generative pipeline, users are fooled 24.7\textperc{} of the time, by the portray module on ground truth sketches even 41.3\textperc{} of the time. We observe a bias of users to assume that 50\textperc{} of images are generated, resulting in a higher rate of misclassified real images for the stronger model. We used 50\textperc{} fake and real images but did not mention this in the task description. For comparison: Isola et al.~\cite{pix2pix} report fooling rates of 18.9\textperc{} and 6.1\textperc{}, however on other modalities.

%%% Local Variables:
%%% mode: latex
%%% TeX-master: "../paper"
%%% End:

%  LocalWords:  discriminative VAE segmentations linearize PCA CDF variational autoencoder CVAE keypoints undistorted unannotated LSUN Chictopia ClothNet misclassified

\section{Conclusion}

In this paper, we developed and analyzed a new approach to generate people with accurate appearance. We find that modern machine learning approaches may sidestep traditional graphics pipeline design and 3D data acquisition. This study is a first step and we anticipate that the results will become better once more data is available for training.

We enhanced the existing Chictopia10K dataset with face annotations and 3D body model fits. With a two-stage model for semantic segmentation prediction in the first, and texture prediction in the second stage, we presented a novel, modular take on generative models of structured, high-resolution images.

In our experiments, we analyzed the realism of the generated data in two ways: by evaluating a segmentation model trained on real data, on our artificial data and by conducting a user study. The segmentation model achieved 85\textperc{} of its segmentation performance of the real data on the artificial, indicating that it `recognized' most parts of the generated images equally well. In the user study, we could in 24.7\textperc{} trick participants into mistaking generated images for real.

With this possibility to generate large amounts of training data at a very low computational and infrastructural cost together with the possibility to condition generated images on pose, shape or color, we see many potential applications for the presented method. We will make data and code available for academic purposes.

\vspace*{-0.5cm}
\paragraph{Acknowledgements} This study is part of the research program of the Bernstein Center for Computational Neuroscience, T\"{u}bingen, funded by the German Federal Ministry of Education and Research (BMBF; FKZ: 01GQ1002). We gratefully acknowledge the support of the NVIDIA Corporation with the donation of a K40 GPU.

%%% Local Variables:
%%% mode: latex
%%% TeX-master: "../paper"
%%% End:

{\small
\bibliographystyle{ieee}
\bibliography{bibliography}

\begin{thebibliography}{10}\itemsep=-1pt

\bibitem{mpiihp}
M.~Andriluka, L.~Pishchulin, P.~Gehler, and B.~Schiele.
\newblock {2D} human pose estimation: New benchmark and state of the art
  analysis.
\newblock In {\em Proc. IEEE Conf. on Computer Vision and Pattern Recognition
  (CVPR)}, June 2014.

\bibitem{SCAPE}
D.~Anguelov, P.~Srinivasan, D.~Koller, S.~Thrun, J.~Rodgers, and J.~Davis.
\newblock {SCAPE}: shape completion and animation of people.
\newblock {\em ACM Trans. Graphics (Proc. SIGGRAPH)}, 24(3):408--416, 2005.

\bibitem{smplify}
F.~Bogo, A.~Kanazawa, C.~Lassner, P.~Gehler, J.~Romero, and M.~J. Black.
\newblock Keep it {SMPL}: Automatic estimation of {3D} human pose and shape
  from a single image.
\newblock In {\em Proc. European Conf. on Computer Vision (ECCV)}, Oct. 2016.

\bibitem{deeplab}
L.-C. Chen, Y.~Yang, J.~Wang, W.~Xu, and A.~L. Yuille.
\newblock Attention to scale: Scale-aware semantic image segmentation.
\newblock In {\em Proc. IEEE Conf. on Computer Vision and Pattern Recognition
  (CVPR)}, pages 3640--3649, 2016.

\bibitem{synth3Dpose}
W.~Chen, H.~Wang, Y.~Li, H.~Su, Z.~Wang, C.~Tu, D.~Lischinski, D.~Cohen-Or, and
  B.~Chen.
\newblock Synthesizing training images for boosting human 3d pose estimation.
\newblock In {\em 3D Vision (3DV)}, 2016.

\bibitem{dai2016shape}
A.~Dai, C.~R. Qi, and M.~Nie{\ss}ner.
\newblock Shape completion using {3D}-encoder-predictor {CNN}s and shape
  synthesis.
\newblock In {\em Proc. IEEE Conf. on Computer Vision and Pattern Recognition
  (CVPR)}, 2017.

\bibitem{deAguiar:2010:SSR}
E.~de~Aguiar, L.~Sigal, A.~Treuille, and J.~K. Hodgins.
\newblock Stable spaces for real-time clothing.
\newblock {\em ACM Trans. Graphics (Proc. SIGGRAPH)}, 29(4):106:1--106:9, July
  2010.

\bibitem{lapgan}
E.~L. Denton, S.~Chintala, R.~Fergus, et~al.
\newblock Deep generative image models using a laplacian pyramid of adversarial
  networks.
\newblock In {\em Advances in Neural Information Processing Systems (NIPS)},
  pages 1486--1494, 2015.

\bibitem{GHFBG07}
R.~Goldenthal, D.~Harmon, R.~Fattal, M.~Bercovier, and E.~Grinspun.
\newblock Efficient simulation of inextensible cloth.
\newblock {\em ACM Trans. Graphics (Proc. SIGGRAPH)}, 26(3):to appear, 2007.

\bibitem{gans}
I.~Goodfellow, J.~Pouget-Abadie, M.~Mirza, B.~Xu, D.~Warde-Farley, S.~Ozair,
  A.~Courville, and Y.~Bengio.
\newblock Generative adversarial nets.
\newblock In {\em Advances in Neural Information Processing Systems (NIPS)},
  pages 2672--2680, 2014.

\bibitem{draw}
K.~Gregor, I.~Danihelka, A.~Graves, D.~J. Rezende, and D.~Wierstra.
\newblock Draw: A recurrent neural network for image generation.
\newblock {\em arXiv preprint arXiv:1502.04623}, 2015.

\bibitem{guan20102d}
P.~Guan, O.~Freifeld, and M.~J. Black.
\newblock A 2d human body model dressed in eigen clothing.
\newblock In {\em Proc. European Conf. on Computer Vision}, pages 285--298.
  Springer, 2010.

\bibitem{DRAPE:2012}
P.~Guan, L.~Reiss, D.~Hirshberg, A.~Weiss, and M.~J. Black.
\newblock {DRAPE: DRessing Any PErson}.
\newblock {\em ACM Trans. Graphics (Proc. SIGGRAPH)}, 31(4):35:1--35:10, July
  2012.

\bibitem{deepercut}
E.~Insafutdinov, L.~Pishchulin, B.~Andres, M.~Andriluka, and B.~Schiele.
\newblock Deepercut: A deeper, stronger, and faster multi-person pose
  estimation model.
\newblock In {\em Proc. European Conf. on Computer Vision (ECCV)}, pages
  34--50. Springer, 2016.

\bibitem{batchnorm}
S.~Ioffe and C.~Szegedy.
\newblock Batch normalization: Accelerating deep network training by reducing
  internal covariate shift.
\newblock {\em arXiv preprint arXiv:1502.03167}, 2015.

\bibitem{h36m_pami}
C.~Ionescu, D.~Papava, V.~Olaru, and C.~Sminchisescu.
\newblock Human3.6m: Large scale datasets and predictive methods for 3d human
  sensing in natural environments.
\newblock {\em IEEE Trans. Pattern Analysis and Machine Intelligence (TPAMI)},
  2014.

\bibitem{pix2pix}
P.~Isola, J.-Y. Zhu, T.~Zhou, and A.~A. Efros.
\newblock Image-to-image translation with conditional adversarial networks.
\newblock In {\em Proc. IEEE Conf. on Computer Vision and Pattern Recognition
  (CVPR)}, 2016.

\bibitem{Jain:2010}
A.~Jain, T.~Thorm\"{a}hlen, H.-P. Seidel, and C.~Theobalt.
\newblock Moviereshape: Tracking and reshaping of humans in videos.
\newblock {\em ACM Trans. Graphics (Proc. SIGGRAPH)}, 29(6):148:1--148:10, Dec.
  2010.

\bibitem{lsp}
S.~Johnson and M.~Everingham.
\newblock Clustered pose and nonlinear appearance models for human pose
  estimation.
\newblock In {\em Proc. British Machine Vision Conf. (BMVC)}, 2010.
\newblock doi:10.5244/C.24.12.

\bibitem{Kavan:2011}
L.~Kavan, D.~Gerszewski, A.~W. Bargteil, and P.-P. Sloan.
\newblock Physics-inspired upsampling for cloth simulation in games.
\newblock {\em ACM Trans. Graphics (Proc. SIGGRAPH)}, 30(4):93:1--93:10, July
  2011.

\bibitem{face_alignment}
V.~Kazemi and J.~Sullivan.
\newblock One millisecond face alignment with an ensemble of regression trees.
\newblock In {\em Proc. IEEE Conf. on Computer Vision and Pattern Recognition
  (CVPR)}, pages 1867--1874, 2014.

\bibitem{Kim:2013:NEP}
D.~Kim, W.~Koh, R.~Narain, K.~Fatahalian, A.~Treuille, and J.~F. O'Brien.
\newblock Near-exhaustive precomputation of secondary cloth effects.
\newblock {\em ACM Trans. Graphics (Proc. SIGGRAPH)}, 32(4):87:1--7, July 2013.

\bibitem{dlib}
D.~E. King.
\newblock Dlib-ml: A machine learning toolkit.
\newblock {\em Journal of Machine Learning Research (JMLR)}, 10:1755--1758,
  2009.

\bibitem{cvae}
D.~P. Kingma, S.~Mohamed, D.~J. Rezende, and M.~Welling.
\newblock Semi-supervised learning with deep generative models.
\newblock In {\em Advances in Neural Information Processing Systems (NIPS)},
  pages 3581--3589, 2014.

\bibitem{vae}
D.~P. Kingma and M.~Welling.
\newblock Auto-encoding variational bayes.
\newblock {\em arXiv preprint arXiv:1312.6114}, 2013.

\bibitem{kulkarni2015deep}
T.~D. Kulkarni, W.~F. Whitney, P.~Kohli, and J.~Tenenbaum.
\newblock Deep convolutional inverse graphics network.
\newblock In {\em Advances in Neural Information Processing Systems (NIPS)},
  pages 2539--2547, 2015.

\bibitem{chictopia10k}
X.~Liang, C.~Xu, X.~Shen, J.~Yang, S.~Liu, J.~Tang, L.~Lin, and S.~Yan.
\newblock Human parsing with contextualized convolutional neural network.
\newblock In {\em Proc. IEEE International Conf. on Computer Vision (ICCV)},
  pages 1386--1394, 2015.

\bibitem{smpl}
M.~Loper, N.~Mahmood, J.~Romero, G.~Pons-Moll, and M.~J. Black.
\newblock Smpl: A skinned multi-person linear model.
\newblock {\em ACM Trans. Graphics (Proc. SIGGRAPH Asia)}, 34(6):248:1--248:16,
  Oct. 2015.

\bibitem{lrelu}
A.~L. Maas, A.~Y. Hannun, and A.~Y. Ng.
\newblock Rectifier nonlinearities improve neural network acoustic models.
\newblock In {\em Proc. International Conf. on Machine Learning (ICML)},
  volume~30, 2013.

\bibitem{Narain:2012:AAR}
R.~Narain, A.~Samii, and J.~F. O'Brien.
\newblock Adaptive anisotropic remeshing for cloth simulation.
\newblock {\em ACM Trans. Graphics (Proc. SIGGRAPH)}, 31(6):147:1--10, Nov.
  2012.

\bibitem{oberweger2015training}
M.~Oberweger, P.~Wohlhart, and V.~Lepetit.
\newblock Training a feedback loop for hand pose estimation.
\newblock In {\em Proc. IEEE International Conf. on Computer Vision (ICCV)},
  pages 3316--3324, 2015.

\bibitem{pixelrnn}
A.~v.~d. Oord, N.~Kalchbrenner, and K.~Kavukcuoglu.
\newblock Pixel recurrent neural networks.
\newblock In {\em Proc. International Conf. on Machine Learning (ICML)}, 2016.

\bibitem{context_encoders}
D.~Pathak, P.~Krahenbuhl, J.~Donahue, T.~Darrell, and A.~A. Efros.
\newblock Context encoders: Feature learning by inpainting.
\newblock In {\em Proc. IEEE Conf. on Computer Vision and Pattern Recognition
  (CVPR)}, pages 2536--2544, 2016.

\bibitem{pishchulin12reshaping}
L.~Pishchulin, A.~Jain, M.~Andriluka, T.~Thorm\"ahlen, and B.~Schiele.
\newblock Articulated people detection and pose estimation: Reshaping the
  future.
\newblock In {\em Proc. IEEE Conf. on Computer Vision and Pattern Recognition
  (CVPR)}, pages 3178--3185. IEEE Press, 2012.

\bibitem{leonid11cvpr}
L.~Pishchulin, A.~Jain, C.~Wojek, M.~Andriluka, T.~Thorm\"ahlen, and
  B.~Schiele.
\newblock Learning people detection models from few training samples.
\newblock In {\em Proc. IEEE Conf. on Computer Vision and Pattern Recognition
  (CVPR)}, 2011.

\bibitem{Pons-Moll:Siggraph2017}
G.~Pons-Moll, S.~Pujades, S.~Hu, and M.~Black.
\newblock {ClothCap}: Seamless {4D} clothing capture and retargeting.
\newblock {\em ACM Trans. Graphics (Proc. SIGGRAPH)}, 36(4), 2017.

\bibitem{PonsTaylorMetricForests2013}
G.~Pons-Moll, J.~Taylor, J.~Shotton, A.~Hertzmann, and A.~Fitzgibbon.
\newblock Metric regression forests for human pose estimation.
\newblock In {\em Proc. British Machine Vision Conf. (BMVC)}. BMVA Press, Sept.
  2013.

\bibitem{rogez16nips}
G.~Rogez and C.~Schmid.
\newblock Mocap-guided data augmentation for 3d pose estimation in the wild.
\newblock In {\em Advances in Neural Information Processing Systems (NIPS)},
  pages 3108--3116, 2016.

\bibitem{rogge2014}
L.~Rogge, F.~Klose, M.~Stengel, M.~Eisemann, and M.~Magnor.
\newblock Garment replacement in monocular video sequences.
\newblock {\em {ACM} Trans. Graphics (Proc. SIGGRAPH)}, 34(1):6:1--6:10, Nov.
  2014.

\bibitem{unet}
O.~Ronneberger, P.~Fischer, and T.~Brox.
\newblock U-net: Convolutional networks for biomedical image segmentation.
\newblock In {\em Proc. International Conf. on Medical Image Computing and
  Computer-Assisted Intervention}, pages 234--241. Springer, 2015.

\bibitem{saito2016photorealistic}
S.~Saito, L.~Wei, L.~Hu, K.~Nagano, and H.~Li.
\newblock Photorealistic facial texture inference using deep neural networks.
\newblock In {\em Proc. IEEE Conf. on Computer Vision and Pattern Recognition
  (CVPR)}, 2017.

\bibitem{shotton2013real}
J.~Shotton, T.~Sharp, A.~Kipman, A.~Fitzgibbon, M.~Finocchio, A.~Blake,
  M.~Cook, and R.~Moore.
\newblock Real-time human pose recognition in parts from single depth images.
\newblock {\em Communications of the ACM}, 56(1):116--124, 2013.

\bibitem{cvae2}
K.~Sohn, H.~Lee, and X.~Yan.
\newblock Learning structured output representation using deep conditional
  generative models.
\newblock In {\em Advances in Neural Information Processing Systems (NIPS)},
  pages 3483--3491, 2015.

\bibitem{Stoll:2010}
C.~Stoll, J.~Gall, E.~de~Aguiar, S.~Thrun, and C.~Theobalt.
\newblock Video-based reconstruction of animatable human characters.
\newblock {\em ACM Trans. Graphics (Proc. SIGGRAPH)}, 29(6):139:1--139:10, Dec.
  2010.

\bibitem{taylor2012vitruvian}
J.~Taylor, J.~Shotton, T.~Sharp, and A.~Fitzgibbon.
\newblock The vitruvian manifold: Inferring dense correspondences for one-shot
  human pose estimation.
\newblock In {\em Proc. IEEE Conf. on Computer Vision and Pattern Recognition
  (CVPR)}, pages 103--110. IEEE, 2012.

\bibitem{pixcnn}
A.~van~den Oord, N.~Kalchbrenner, L.~Espeholt, O.~Vinyals, A.~Graves, et~al.
\newblock Conditional image generation with pixelcnn decoders.
\newblock In {\em Advances in Neural Information Processing Systems (NIPS)},
  pages 4790--4798, 2016.

\bibitem{learning_from_synthetic}
G.~Varol, J.~Romero, X.~Martin, N.~Mahmood, M.~Black, I.~Laptev, and C.~Schmid.
\newblock Learning from synthetic humans.
\newblock In {\em Proc. IEEE Conf. on Computer Vision and Pattern Recognition
  (CVPR)}, 2017.

\bibitem{virtual_humans}
D.~V{\'a}zquez, A.~M. Lopez, J.~Marin, D.~Ponsa, and D.~Geronimo.
\newblock Virtual and real world adaptation for pedestrian detection.
\newblock {\em IEEE Trans. Pattern Analysis and Machine Intelligence (TPAMI)},
  36(4):797--809, 2014.

\bibitem{Wang:SIGGRAPH:2011}
H.~Wang, J.~F. O'Brien, and R.~Ramamoorthi.
\newblock Data-driven elastic models for cloth: Modeling and measurement.
\newblock {\em ACM Trans. Graphics (Proc. SIGGRAPH)}, 30(4):71:1--11, July
  2011.

\bibitem{Xu:2011:VCC}
F.~Xu, Y.~Liu, C.~Stoll, J.~Tompkin, G.~Bharaj, Q.~Dai, H.-P. Seidel, J.~Kautz,
  and C.~Theobalt.
\newblock Video-based characters: Creating new human performances from a
  multi-view video database.
\newblock {\em ACM Trans. Graphics (Proc. SIGGRAPH}, 30(4):32:1--32:10, July
  2011.

\bibitem{attribute2image}
X.~Yan, J.~Yang, K.~Sohn, and H.~Lee.
\newblock Attribute2image: Conditional image generation from visual attributes.
\newblock In {\em Proc. European Conf. on Computer Vision (ECCV)}, pages
  776--791. Springer, 2016.

\bibitem{yang2016high}
C.~Yang, X.~Lu, Z.~Lin, E.~Shechtman, O.~Wang, and H.~Li.
\newblock High-resolution image inpainting using multi-scale neural patch
  synthesis.
\newblock In {\em Proc. IEEE Conf. on Computer Vision and Pattern Recognition
  (CVPR)}, 2017.

\bibitem{lsun}
F.~Yu, Y.~Zhang, S.~Song, A.~Seff, and J.~Xiao.
\newblock Lsun: Construction of a large-scale image dataset using deep learning
  with humans in the loop.
\newblock {\em arXiv preprint arXiv:1506.03365}, 2015.

\bibitem{deconv}
M.~D. Zeiler, G.~W. Taylor, and R.~Fergus.
\newblock Adaptive deconvolutional networks for mid and high level feature
  learning.
\newblock In {\em Proc. IEEE International Conf. on Computer Vision (ICCV)},
  pages 2018--2025. IEEE, 2011.

\bibitem{shape_under_cloth:CVPR17}
C.~Zhang, S.~Pujades, M.~Black, and G.~Pons-Moll.
\newblock Detailed, accurate, human shape estimation from clothed {3D} scan
  sequences.
\newblock In {\em Proc. IEEE Conf. on Computer Vision and Pattern Recognition
  (CVPR)}, 2017.

\bibitem{zhou2010parametric}
S.~Zhou, H.~Fu, L.~Liu, D.~Cohen-Or, and X.~Han.
\newblock Parametric reshaping of human bodies in images.
\newblock {\em ACM Trans. Graphics (Proc. SIGGRAPH)}, 29(4):126, 2010.

\end{thebibliography}
}

\end{document}